%% file: main.tex
\documentclass[review]{fcs}
\usepackage{fontawesome}
\usepackage{multicol}
\usepackage{balance}
\usepackage{float} 
\usepackage{amsmath}
\usepackage{amsthm}
\usepackage{ragged2e} 
\usepackage{url}
\usepackage{booktabs,makecell, multirow, tabularx}
\usepackage{colortbl}
\usepackage{algorithmic}
\usepackage{adjustbox}
\usepackage[ruled,linesnumbered,vlined]{algorithm2e}
\usepackage{enumitem}
\usepackage{verbatim}
\usepackage{CJKutf8}

\newcommand{\Mat}[1]{\mathbf{#1}}

\newcommand{\ie}{\textit{i.e., }}
\newcommand{\eg}{\textit{e.g., }}

\newcommand{\cf}{\textit{cf. }}

\renewcommand{\thefootnote}{\fnsymbol{footnote}}
\definecolor{mygray}{rgb}{0.9, 0.9, 0.9}
\definecolor{gray}{rgb}{0.5, 0.5, 0.5}
\title{Invariant Graph Learning Meets Information Bottleneck for Out-of-Distribution
Generalization}
\author[1]{Wenyu Mao}
\author[2]{Jiancan Wu (\Letter)}
\author[2]{Haoyang Liu}
\author[3]{Yongduo Sui}
\author[4]{Xiang Wang}
\address[1]{School of Cyber Science and Technology, University of Science and Technology of China, Hefei
230026, China}
\address[2]{School of Infomation Science and Technology, University of Science and Technology of China, Hefei
230026, China}
\address[3]{Tencent, Shenzhen
518054, China}
\address[4]{School of Artificial Intelligence and Data Science, University of Science and Technology of China, Hefei
230026, China}
\fcssetup{
  received       = {January 15, 2025},
  accepted       = {January 15, 2025},
  corr-email     = {wujcan@gmail.com},
}
\begin{abstract}
Graph out-of-distribution (OOD) generalization remains a major challenge in graph learning since graph neural networks (GNNs) often suffer from severe performance degradation under distribution shifts. Invariant learning, aiming to extract invariant features across varied distributions, has recently emerged as a promising approach for OOD generalization. 
Despite the great success of invariant learning in OOD problems for Euclidean data (\ie images), the exploration within graph data remains constrained by the complex nature of graphs. The invariant features at both the attribute and structural levels, combined with the absence of prior knowledge regarding environmental factors, make the invariance and sufficiency conditions of invariant learning hard to satisfy on graph data. Existing studies, such as data augmentation or causal intervention, either suffer from disruptions to invariance during the graph manipulation process or face reliability issues due to a lack of supervised signals for causal parts.
In this work, we propose a novel framework, called Invariant Graph Learning based on Information bottleneck theory (InfoIGL), to extract the invariant features of graphs and enhance models' generalization ability to unseen distributions. Specifically, InfoIGL introduces a redundancy filter to compress task-irrelevant information related to environmental factors. Cooperating with our designed multi-level contrastive learning, we maximize the mutual information among graphs of the same class in the downstream classification tasks, preserving invariant features for prediction to a great extent. An appealing feature of InfoIGL is its strong generalization ability without depending on supervised signal of invariance. 
Experiments on both synthetic and real-world datasets demonstrate that our method achieves state-of-the-art performance under OOD generalization for graph classification tasks. The source code is available at \url{https://github.com/maowenyu-11/InfoIGL}.
\end{abstract}
\keywords{Graph OOD, Contrastive Learning, Information Bottleneck Theory, Invariant Learning}

\begin{document}

\input{1_introduction.tex}

\input{2_related_work.tex}

\input{3_preliminary}
\input{4_methodology.tex}
\input{5_experiments.tex}

\input{6_conclusion.tex}
\input{7_appendix}
\bibliographystyle{fcs}
\bibliography{main}

\end{document}

%% file: 1_introduction.tex
\section{Introduction}
\label{intro}
Graphs are ubiquitous in the real world, appearing as chemical molecules, recommender systems \cite{FCS_wu,RPP_TOIS,mao2025distinguishedquantizedguidancediffusionbased,tois_Wu}, and knowledge graphs, to name a few examples.
In recent years, graph neural networks (GNNs)~\cite{WWW_GIF,xu2018powerful} have emerged as a potent representation learning technique for analyzing and making predictions on graphs. Despite significant advancements, most existing GNN approaches rely heavily on the i.i.d. assumption that the distribution of test data is independently and identically distributed to the training data. Such an assumption, however, seldom holds in practice due to environmental asynchrony during data collection, leading to distribution shifts between the training and testing data \cite{tkdd_ood}. In these situations, GNN suffers from severe performance degradation, making graph OOD generalization a significant challenge in graph learning.

\begin{figure}[t]%靠文字内容的右侧

\centering
\includegraphics[width=0.98\linewidth]{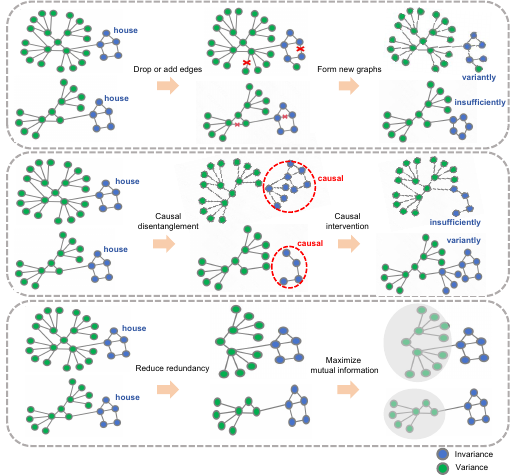}
% \vspace{-4mm}
\caption{The comparison of the two branches of existing methods and our method. {The upper is data manipulation, which edits nodes or edges and is prone to destroying the invariant parts. 
The causal disentanglement approach in the middle subfigure separates causal subgraphs and intervenes across various environmental conditions (green patterns), which may fail to identify the causal parts accurately. 
In contrast, our method in the lower subfigure first removes most of the redundancy to avoid distraction for invariance discovery, facilitating higher identifying accuracy.
Then it preserves sufficient predictive information (house) in invariance by maximizing the mutual information, without the use of supervised labels for invariance.} }
\vspace{-0.5cm}
\label{fig:fig1}

\end{figure}
Identifying graph features that remain invariant across distribution shifts~\cite{wang2022unified, DIR} is paramount in overcoming the graph OOD problem. 
Invariant learning assumes that invariant features sufficiently determine the label while spurious features are affected by task-irrelation environmental factors cross distributions~\cite{arjovsky2019invariant}. Two critical conditions of graph invariance need to be guaranteed in this process~\cite{wu2022handling,liu2022graph,DIR,li2022learning}. 
\begin{itemize}
    \item
    {Invariance condition: The graph invariance should exclude spurious features related to environmental factors and maintain robustness across diverse distributions.
    \label{inv}}
 \item 
   {Sufficiency condition: The graph invariance must remain intact and contain sufficient information to predict labels accurately.
   \label{suf}}
\end{itemize}
Existing studies of invariant graph learning mainly focus on two lines to solve the graph OOD problem: graph manipulation and causal disentanglement approaches. 
On the one hand, \textit{graph manipulation approaches}~\cite{wang2021mixup} typically first generate diverse augmented data (\eg adding or removing nodes and edges) to promote the diversity of training distributions, then learn representations consistent on these distributions.
This line of research often leads to inappropriate augmentations and destroys the {\textbf{invariance} accidentally}. 
{As shown in the upper subfigure in Figure \ref{fig:fig1}, data augmentation disrupts the invariant parts (the blue patterns) of the original graphs. Consequently, the newly generated graphs are unable to maintain \textbf{invariance} or \textbf{sufficiently} predict the labels ``house''.}
On the other hand, \textit{causal disentanglement methods}~\cite{fan2021generalizing,li2021ood, DIR,sui2022causal,liu2022graph} {aim} to mitigate confounding effects of environments and separate the underlying causal subgraphs according to causal intervention theory~\cite{pearl2014interpretation}. {Identifying and separating the causal parts from the non-causal parts is challenging due to two factors. (1) The lack of supervised signals. It is almost impossible to label the causal parts in a graph manually, making the training process for identifying causal subgraphs intractable. (2) The high complexity of the graph. The causal parts
may occur in terms of both features (color or properties) and structures (nodes and edges), posing great difficulties in separating the causal parts accurately. As depicted in the middle subfigure of Figure \ref{fig:fig1}, the separated blue pattern is not \textbf{invariant} across different green patterns, nor is it \textbf{sufficient} for accurately predicting the label ``house''.}
Both lines are limited in identifying graph invariance that satisfies the two conditions.

To circumvent the aforementioned limitations, we propose a novel framework \textbf{InfoIGL} for invariant graph learning inspired by information bottleneck theory~\cite{saxe2019information}, to meet the invariance and sufficiency conditions. The goal is to compress redundant information in graphs to exclude spurious features while maximizing task-relevant information for prediction as invariance, {as demonstrated in Figure~\ref{fig:fig1}}. Specifically, to exclude spurious features, we implement a redundancy filter that assigns minimal invariance scores to redundant information. To preserve sufficient predictive information in invariance, we convert the goal into maximizing the mutual information of graphs from the same class. {To fully realize the mutual information maximization, we employ multi-level contrastive learning~\cite{yue2021prototypical,wang2022cross} (\ie{semantic- and instance-level}), without relying on supervised signals of the invariance. The semantic level focuses on global category semantics that remain stable across complex graphs, while the instance level emphasizes detailed information from local graph samples. This multi-level contrastive learning strategy facilitates a more comprehensive maximization of intra-class mutual information, enhancing the robustness of invariance and satisfying the sufficiency condition.}

To prevent contrastive loss from target collapse, where it fails to distinguish between positive and negative samples~\cite{yao2022pcl}, InfoIGL strengthens instance level contrastive learning with instance constraint and hard negative mining techniques. The outperforming results of extensive experiments on both synthetic and real-world datasets demonstrate the effectiveness of InfoIGL. 
Our main contributions are summarized as follows.
\begin{itemize}
\item To address the graph OOD problem in classification tasks, we propose a novel framework called InfoIGL inspired by information bottleneck theory, which maximizes mutual information of graphs as invariance after compressing the redundant information.

\item We incorporate multi-level contrastive learning from both semantic and instance levels to maximize mutual information of graphs in the same class without supervised signals for invariance.

\item We conduct extensive experiments on diverse benchmark datasets to demonstrate the effectiveness of our proposed framework InfoIGL.

\end{itemize}

%% file: 2_related_work.tex
\section{Related Work}

\noindent\textbf{Invariant learning for graph OOD.} 
Invariant learning~\cite{zhao2019learning,rosenfeld2020risks,yang2020learning,wang2022unified} has been extensively explored to improve generalization performance on graph OOD scenarios, which learns robust representations withstanding distribution shifts. Growing research is concentrating on applying invariant learning strategies to tackle the problem of graph OOD generalization, such as optimization methods~\cite{arjovsky2019invariant}, causal learning~\cite{DIR,sui2022causal,tkdd_ood}, stable learning~\cite{fan2021generalizing,li2021ood}, and data manipulation~\cite{miao2022interpretable,rong2019dropedge}.
Optimization methods design optimization objectives to make the model robust to the shifts in data distribution~\cite{arjovsky2019invariant,krueger2021out,sagawa2019distributionally,kong2022robust,federici2021information}. Causal learning utilizes causal intervention to capture causal features for prediction and ignore non-causal features~\cite{DIR,sui2022causal,liu2022graph,fan2022debiasing,sui2024unleashing,yang2022learning,gui2024joint,zhuang2024learning}. Stable learning leverages sample reweighting to eliminate spurious correlation and extract stable features across different environments~\cite{fan2021generalizing,li2021ood}. Data manipulation~\cite{miao2022interpretable,wang2021mixup,han2022g,li2022learning,chen2024does} such as dropEdge~\cite{rong2019dropedge} randomly drops the edges of graphs to increase the diversity of data distribution. 
However, many of them overlook the intricacies of graph data and are limited by the lack of theoretical guarantees for direct application.

\noindent\textbf{The information bottleneck theory and boundaries of mutual information.} Existing research\cite{tishby2000information, fang2024regularization} has delineated the paradigm of acquiring a robust representation through the lens of the information bottleneck theory~\cite{saxe2019information}. This theoretical framework endeavors to optimize the mutual information between the derived representation and predictive outcomes while concurrently minimizing the mutual information between the representation and the original input. This selective process aims at preserving solely the salient information pertinent to the underlying task at hand. Within the context of distribution shifts~\cite{ye2022ood}, the pursuit of a robust representation via information bottleneck theory aligns with the extraction of invariant features~\cite{du2020learning,
ahuja2021invariance,
li2022invariant}. Nonetheless, the direct computation of mutual information encounters formidable obstacles when dealing with high-dimensional continuous variables of graphs. Given that the precise calculation of mutual information often proves extraneous to the core objectives, models are tailored towards delineating the boundaries of mutual information~\cite{poole2019variational} for optimization purposes. Poole \cite{poole2019variational} has expounded upon the intricate interplay between mutual information and its assorted lower bounds, wherein the contrastive loss mechanism emerges as a prevalent methodological choice.

\noindent\textbf{Contrastive learning and OOD generalization.} 
Contrastive learning~\cite{he2020momentum,chen2020simple,khosla2020supervised} has achieved success in aligning representations by pulling together positive pairs and pushing apart negative pairs. Minimizing the contrastive loss can maximize the mutual information between positive pairs while maximizing that between negative pairs. Such a strategy ensures that the mutual information of inputs from the same class encapsulates information relevant to the target, as highlighted in ~\cite{tian2020makes,hassani2020contrastive}, which is stable in the face of distributional shifts as the invariance. This approach thereby establishes crucial invariance necessary for prediction.
Recently, the success of leveraging contrastive learning for domain generalization tasks~\cite{huang2021towards,yao2022pcl,zhang2022correct} in the area of computer vision attracts researchers' attention in addressing OOD problems in the graph scenarios. For instance, CIGA~\cite{chenlearning} applies contrastive learning after decomposing the graph causality which contains the most information on labels.
Unlike previous work, we utilize contrastive learning on the features after reducing redundancy to avoid including spurious features in the invariance. We encourage intra-class compactness and inter-class discrimination~\cite{wang2022cross} with class labels to maximize the predictive information. Additionally, {to enhance the robustness of graph invariance and satisfy the sufficiency condition}, we conduct contrastive learning from both semantic and instance levels to fully maximize the mutual information between graphs.

%% file: 3_preliminary.tex
\section{Preliminaries}
\subsection{Problem Formulation of graph OOD}

Let $\mathbb{G}$ and $\mathbb{Y}$ be the sample space and label space, respectively. 
We denote a sample graph by $G \in \mathbb{G}$ with the adjacent matrix $\Mat{A}$ and node feature matrix $\Mat{X}$. The bold $\Mat{G}$ and $\Mat{Y}$ are random variables for graphs and labels.
Suppose that $\mathcal{D}_{\mathrm{tr}}=\{(G_i^e, Y_i^e)\}_{e \in \mathcal{E}_{\mathrm{tr}}}$ and $\mathcal{D}_{\mathrm{te}}=\{(G_i^e, Y_i^e)\}_{e \in \mathcal{E}_{\mathrm{te}}}$ be the training and testing dataset, where $e$ donates the environment from training environment sets $\mathcal{E}_{\mathrm{tr}}$ and testing environment sets $\mathcal{E}_{\mathrm{te}}$. The training and testing distributions are often inconsistent due to different environmental factors, \ie $P(\Mat{G}^e, \Mat{Y}^e|e=e_1) \neq P(\Mat{G}^e, \Mat{Y}^e|e=e_2)$ with $e_1 \in \mathcal{E}_{\mathrm{tr}}$ and $e_2 \in \mathcal{E}_{\mathrm{te}}$. 
A graph predictor $f=\theta \circ \Phi: \mathbb{G} \to \mathbb{Y}$ maps the input graph $G$ to the corresponding label $Y \in \mathbb{Y}$, which can be decomposed into a graph encoder $\Phi$ and a classifier $\theta$.
Consequently, the aim of generalization for graph OOD is to train models $f=\theta \circ \Phi$ with $\mathcal{D}_{tr}$
to generalize well on unseen distributions $\mathcal{D}_{te}$, which can be formulated as follows:
\begin{gather}
\underset{f}{\mathrm{min}} \ \underset{e\in \mathcal{E}_{\mathrm{te}}}{\mathrm{max}} \,\mathbb{E}_{G^e,Y^e \in \mathcal{D}_{te}} [\mathcal{L}(Y^e,f(G^e)]
\label{eq: obj}
\end{gather}
where $\mathcal{L}(\cdot, \cdot )$ is the loss function between the ground truth and predicted labels.

% \begin{figure}%靠文字内容的右侧
% % \vspace{-3mm}
% \centering
% \includegraphics[width=\linewidth]{Figs/fig2.png}
% % \vspace{-4mm}
% \caption{Task decomposition. The invariant features are the shared information of graphs which is stable for the prediction of $Y$. Task1: Compressing the redundant information of $g$ to $\Phi(g)$ to minimize $I(g; \Phi(g))$; Task2: Maximizing the mutual information between $\Phi(g_1)$ and $\Phi(g_2)$ to retain the useful information for prediction merely. }
% \label{fig:fig2}
% % \vspace{-4mm}
% \end{figure}

\subsection{Invariant Learning and Information Bottleneck Theory}
\label{sec: inv-info}
The objective of Equation~\ref{eq: obj} is hard to optimize since the prior knowledge for test environments is not available during the training process. Recent
studies~\cite{chenlearning, arjovsky2019invariant, wu2022handling,Auxiliaryood} focus on invariant learning to solve the problem of OOD. The basic idea is that the variables $z$ in the latent space $Z$ can be partitioned into invariant and spurious parts.
They define the invariant features $\Mat{z}_{inv}$ to be those sufficient for the prediction task~\cite{yang2024invariant}  while the spurious features $\Mat{z}_{spu}$ be the task-irrelevant~\cite{sui2022causal, DIR} ones related to the environment $e$. We formulate it as below:
\begin{equation}
\Mat{Y} \perp e \, | \,\Mat{z}_{inv},\quad I(\Mat{Y}; \Mat{z}_{spu} |\Mat{z}_{inv})=0,
\label{con: inv}
\end{equation}
\begin{equation}
 I(\Mat{Y}; \Mat{z}_{inv})=I(\Mat{Y}; \Mat{G}) 
 \label{con:suf}
\end{equation}
% \end{array}
% \end{equation}
where $I(\cdot;\cdot)$ represents the mutual information between two variables, $\perp$ denotes independence, Equation~\ref{con: inv} denotes the invariance condition while Equation~\ref{con:suf} denotes the sufficiency condition.
 Different from existing work based on data manipulation~\cite{miao2022interpretable,rong2019dropedge,wang2021mixup,han2022g} and causal disentanglement~\cite{fan2021generalizing,li2021ood, DIR,sui2022causal,liu2022graph}, we take inspirations from the information bottleneck theory~\cite{tishby2000information} for invariant learning, where the goal for information bottleneck is define by: 
\begin{gather}
R_{IB}(\theta)=I(\Phi(\Mat{G}); \Mat{Y})-\beta I(\Phi(\Mat{G}); \Mat{G}).
\label{rib}
\end{gather}
For invariant learning, the goal is to optimize the encoder $\Phi(\Mat{G})$ by minimizing the mutual information $I(\Phi(\Mat{G}); \Mat{G})$ to reduce the redundancy from $\Mat{z}_{spu}$ while maximizing $I(\Phi(\Mat{G}); \Mat{Y})$ to satisfy the sufficiency condition~\cite{wu2022handling} for prediction.
% \begin{gather}
% I(\Mat{Y}; \Mat{z}_{inv})=\max I(\Mat{Y;\Phi(\Mat{G})},\quad
% I(\Mat{G}; \Mat{z}_{inv})=\min I(\Mat{G}; \Phi(\Mat{G}))
% \end{gather}
Therefore, the encoder $\Phi(\Mat{G})$ is trained to approximate the invariance features, \ie{$\Mat{z}_{inv}\approx\Phi(\Mat{\Mat{G}})$}.
% The encoder $\Phi(\Mat{G})$ aims to obtain the invariant graph features $\Mat{z}_{inv}$, \ie{$\Mat{z}_{inv}\approx\Phi(\Mat{\Mat{G}})$}. 

\subsection{Multi-level Contrastive Learning}
% Contrastive learning facilitates invariant learning in the absence of supervised signals for invariance. 
\label{sec: multi-level}
In practice, the supervised signals for invariance are intractable to obtain. To facilitate invariant learning, contrastive learning can serve as a practical approximation to identify invariance. {To satisfy the sufficiency condition, multi-level contrastive learning from both instance and semantic levels can fully capture the invariant features from complex graph data and enhance the robustness of graph invariance}. Specifically, instance-level contrastive learning aims to maintain pairwise similarity within instances shared with the same labels in the classification task~\cite{xu2021self, zhang2022leverage,khosla2020supervised}. In light of this, the instances with the same label serve as positive samples and are pushed closer in the latent space.
In contrast, instances with different labels are negative samples and separated apart in the latent space.
Consequently, the objective is 
\begin{gather}
 \mathcal{L}_{\mathrm{ins}}=\mathbb{E}\left[-\mathrm{log}\frac{\mathrm{exp}(\Mat{z} \cdot \Mat{z}_+/\tau)}{\mathrm{exp}(\Mat{z} \cdot \Mat{z}_+/ \tau)+\sum\mathrm{exp}(\Mat{z}\cdot \Mat{z_{-}}/\tau)}\right]
\end{gather}
where $\Mat{z}$, $\Mat{z}_+$ and $\Mat{z_{-}}$ denote the features of input graphs and their corresponding positive and negative instances, respectively.

Semantics are category centers that represent semantic features for each category~\cite{yao2022pcl}, obtained through methods such as clustering. Semantic-level contrastive learning aims to compactly embed instances around the corresponding semantic while also refining the semantic to better represent the class by minimizing the loss:
\begin{gather}
 \mathcal{L}_{\mathrm{sem}}=\mathbb{E}\left[-\mathrm{log}\frac{\mathrm{exp}(\Mat{z}\cdot \Mat{w}_{+}/ \tau)}{\mathrm{exp}(\Mat{z}\cdot \Mat{w}_{+}/\tau)+\sum\mathrm{exp}(\Mat{z}\cdot \Mat{w}_{-}/\tau)}\right]
\end{gather}
where $\Mat{z}$, $\Mat{w}_+$ and $\Mat{w_{-}}$ denote the features of instances and their corresponding positive and negative semantics, respectively. Semantics that are the same class as the instances $\Mat{z}$ are positive semantic $\Mat{w}_+$ while negative semantics $\Mat{w_{-}}$ are those from other classes.

{Instance-level contrastive methods may overlook global category features and exhibit instability due to the complex nature of different graphs. Conversely, semantic-level contrastive learning may sacrifice the exploration of detailed local features in favor of invariance}. Incorporating multi-level contrastive learning~\cite{xu2021self,zhang2022leverage} can extract invariant features {\textbf{to the greatest extent}}, {enhancing the robustness of invariance and satisfying the sufficiency condition.}

%% file: 4_methodology.tex
\section{Methodology}
To satisfy the invariance and sufficiency conditions of invariant learning, we propose to extract invariant graph representations from the perspective of information bottleneck theory. In this section, we first adapt the goals of information bottleneck theory to invariant learning. Then we introduce a novel framework, called InfoIGL according to it, thus solving the graph OOD problem.

% Specifically, InfoIGL first utilizes a redundancy filter to filter out spucious features. Then it maximizes the mutual information of graphs from the same class by multi-level contrastive learning. Finally, InfoIGL transfers the invariant representation to the downstream task---graph classification.
% Maximizing mutual information can optimize the step of compression redundancy, thus satisfying the two conditions for invariance. The framework is illustrated in Figure~\ref{fig:model}. 

\subsection{Rethinking Information Bottleneck Theory for Invariant Graph Learning }
\label{sec:4.1}

According to the information bottleneck theory for invariant learning in Section~\ref{sec: inv-info}, we should minimize the mutual information
 $I(\Phi(\Mat{G}); \Mat{G})$ to the lower bound $I(\Mat{z}_{inv}; \Mat{G})$ while maximizing $I(\Phi(\Mat{G}); \Mat{Y})$ to the upper bound $I(\Mat{z}_{inv}; \Mat{Y})$.
However, since the supervised signals for invariance are unrealistic to obtain in practice, we thus return to a surrogate objective for training the encoder $\Phi(\cdot)$ based on the theorem proposed in CNC~\cite{zhang2022correct} and CIGA~\cite{chenlearning}: maximizing the mutual information between samples from the same class can approximate maximizing the predictive information for invariance. We formulate it as below: 

\begin{gather}
\max I(\Phi(\Mat{G}) ; \Mat{Y}) \rightarrow \underset{\widehat{G}, \widetilde{G}, Y \in D_{tr}}{\max } I(\Phi(\widehat{G}) ; \Phi(\widetilde{G})\mid Y)
\label{eq: max_goal}
\end{gather}
\begin{figure*}[t]
    \centering
    \includegraphics[width=1\linewidth]{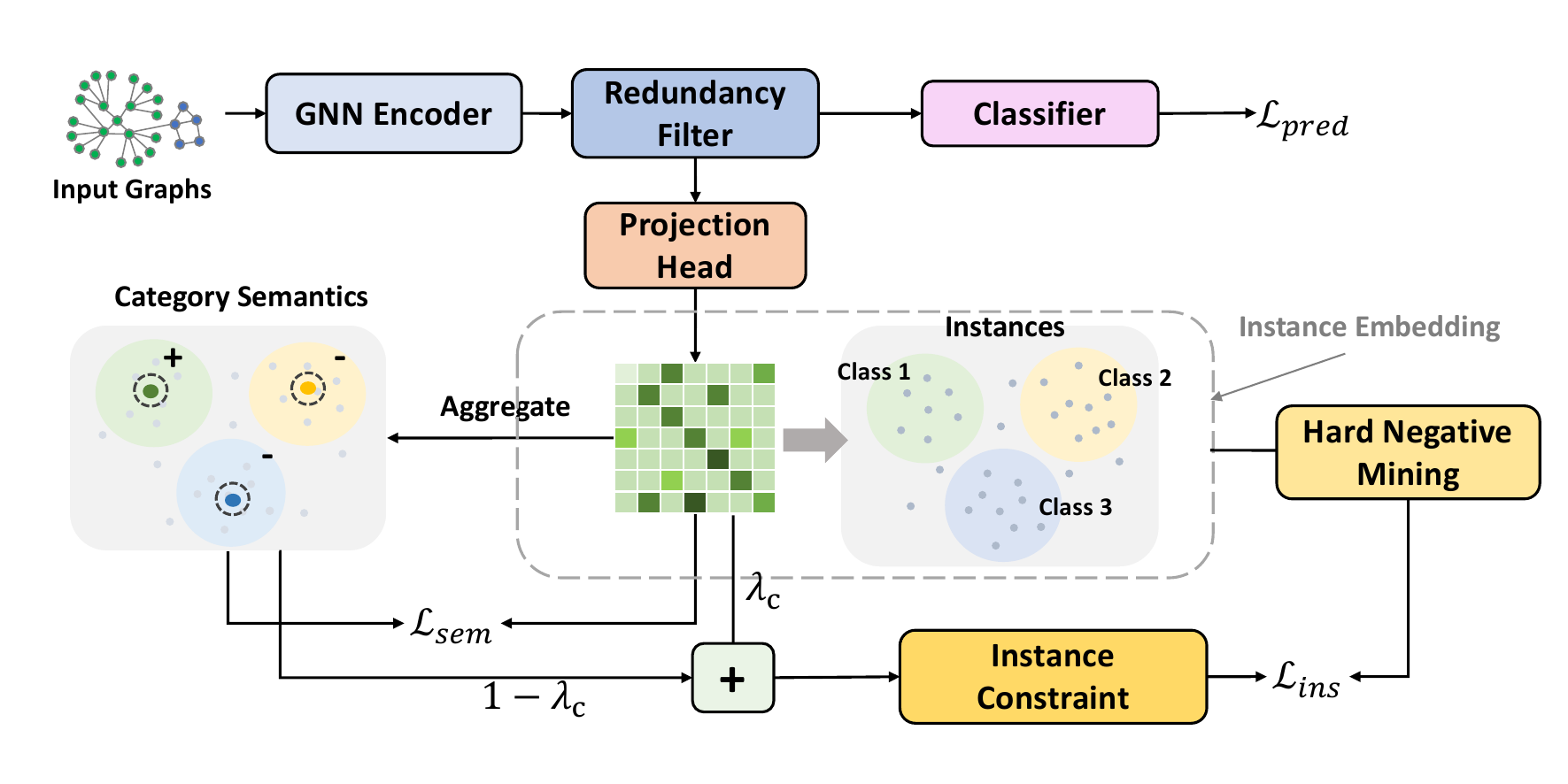}
    % \vspace{-2mm}
    \caption{The overview of proposed InfoIGL framework. 
    The training graphs are fed into the GNN encoder and attention mechanism~\cite{sui2022causal,brody2021attentive}. After being projected to another space, instance embeddings are aggregated to semantics. Then semantic-level and instance-level contrastive learning are optimized jointly, along with instance constraint and hard negative mining to avoid model collapse.
    }
    \label{fig:model}
    % \vspace{-2mm}
\end{figure*}
 
To solve the above optimization problem of maximizing $I(\Phi(\widehat{G}); \Phi(\widetilde{G})\mid Y)$, we resort to the mutual information boundary~\cite{poole2019variational,mutualesti} as is shown below.

\noindent
\textbf{Mutual information boundary:}
Given two variables $X, Y$, we derive the lower bound of the mutual information $I(X;Y)$ by InfoNCE loss:
\begin{equation}
\begin{split}
I(X;Y)\geq \mathrm{E}\left[\frac{1}{K}\sum \limits_{i=1}^{K}\mathrm{log}\frac{\mathrm{exp}{(\phi(x_{i},y_{i}))}}{\frac{1}{K}\sum_{j=1}^{K}\mathrm{exp}{(\phi(x_{i},y_{j})})}\right]\triangleq I_{NCE}
\label{eq:mutual_boundary}
\end{split}
\end{equation}
where $x_i$ and $y_i$ denote a positive pair sampled from the joint distribution $P(X, Y)$, while $x_i$ and $y_j$ form a negative pair sampled from the product of marginal distributions $P(X)P(Y)$, $\phi$ is the similarity function. This inspires us to leverage contrastive learning to maximize mutual information $I(\Phi(\widehat{G}); \Phi(\widetilde{G})\mid Y)$.
 % As defined in Equation~\eqref{eq:mutual_boundary}, maximizing the mutual information equals minimizing the InfoNCE loss~\cite{oord2018representation}, which inspires us to leverage contrastive learning to accomplish Task II.\hfill $\square$
\subsection{Implementation of InfoIGL}
Based on Equation~\ref{rib}, 
InfoIGL first optimizes the encoder with a redundancy filter which minimizes the mutual information $I(\Phi(\Mat{G}); \Mat{G})$ by filtering out spurious features $\Mat{z}_{spu}$. According to Equation~\ref{eq: max_goal} and~\ref{eq:mutual_boundary}, it maximizes the mutual information of graphs from the same class by multi-level contrastive learning as introduced in Section~\ref{sec: multi-level}. Finally, InfoIGL transfers the invariant representation to the downstream task---graph classification.
Multi-level contrastive learning can further optimize the redundancy filter, thus satisfying the invariance and sufficiency conditions for invariant learning. The framework is illustrated in Figure~\ref{fig:model}. 
\subsubsection{Reducing Redundancy} We implement a redundancy filter to remove task-irrelevant information from graphs and minimize the mutual information $I(\Phi(\Mat{G}); \Mat{G})$ in information bottleneck theory. The filter assigns minimal invariance scores for spurious features, which can be realized through attention mechanism~\cite{sui2022causal,brody2021attentive}. 
Before applying the redundancy filter, we obtain the representations for graph nodes with GNNs first. We can build our framework on any GNN backbones, and here we take GIN~\cite{xu2018powerful} as an example, the node update module is defined as follows:
\begin{gather}
\Mat{h}_v^{(k)}=\mathrm{MLP}^k((1+\epsilon^{(k)})\cdot \Mat{h}_v^{(k-1)}+\sum_{u \in N(v)} \Mat{h}_u^{(k-1)})
\label{eq:GIN_update}
\end{gather}
where $\mathrm{MLP}(\cdot)$ stands for multi-layer perceptron, $\epsilon$ is a learnable parameter, $\Mat{h}_v$ and $\Mat{h}_u$ separately denote the representations of nodes $v$ and $u$, $N(v)$ denotes the neighbour nodes of $v$. Then we assign invariance scores to node $v$ and edge $(u,v)$ through the attention mechanism, which can be obtained as follows:
\begin{gather}
\alpha_v=\mathrm{softmax}(\frac{Q_vK_v^\top}{\sqrt{d_k}})V_v\\
\alpha_{uv}=\mathrm{softmax}(\mathrm{LeakyReLU}(\mathrm{MLP}(\Mat{h}_u||\Mat{h}_v)))
\label{eq:att}
\end{gather}
where $ Q_v=\Mat{h}_v\cdot \Mat{W}^Q,K_v=\Mat{h}_v\cdot \Mat{W}^K,V_v=\Mat{h}_v\cdot \Mat{W}^V$, $\Mat{W}^Q,\Mat{W}^K,\Mat{W}^V$ are trainable parameter matrices, $d_k$ is $K_v$'s dimension,  $||$ is the concatenation operation.
After assigning invariance scores for nodes and edges, we encode graphs into graph-level representations $\Mat{h}_G$:
\begin{gather}
\Mat{h}_v'=\alpha_v\cdot \Mat{h}_v, \Mat{h}_{vu}'=\alpha_{uv}\cdot \Mat{h}_{uv}\\
\Mat{h}_G=\mathrm{READOUT}(\Mat{H}_v,\Mat{H}_{uv})
\label{eq:readout}
\end{gather}
where $\Mat{h}_v,\Mat{h}_{uv}$ are the node and edge embedding obtained from GIN. We denote $\Mat{H}_v=[...,\Mat{h}_v',...]_{v\in \mathcal{V}}^\top$, $\Mat{H}_{uv}=[...,\Mat{h}_{uv}',...]_{u,v\in \mathcal{V}}^\top$, $\mathcal{V}$ is the node set. 
Through the redundancy filter, InfoIGL can filter out spurious features in graphs and minimize 
 $I(\Phi(\Mat{G}); \Mat{G})$ in Equation~\ref{rib}.

\noindent
\subsubsection{Maximizing Predictive Information}
\label{task2_method}
To maximize predictive information $I(\Phi(\Mat{G}); \Mat{Y})$ and further optimize the redundancy filter, contrastive learning with supervised class labels provides a practical solution to maximize the mutual information between graphs from the same class based on Equation~\ref{eq: max_goal} and~\ref{eq:mutual_boundary}. {To enhance the robustness of graph invariance and satisfy the sufficiency condition}, we implement multi-level contrastive learning at both the semantic and instance levels according to section~\ref{sec: multi-level}. 
% However, traditional contrastive-based methods may not effectively work since complex instance-wise relationships may impede model generalization\cite{yao2022pcl,xu2021self,zhang2022leverage}. To fully unleash the advantages of contrastive learning, we jointly optimize semantic-wise and instance-wise contrastive losses after projecting the embedding to another space with a projection head, which can promote both inter-class separation and intra-class compactness. 

\noindent\textbf{Projection Head. }
\label{sec:projection}
% A projection head is a small network that maps the embedding to another space where further contrastive learning is applied ~\cite{yao2022pcl,gupta2022understanding}, which can prevent the conflicts between reducing representation entropy and strengthening contrastive uniformity according to the Proposition~\ref{pro:proj}.
% \begin{myPro}
% \label{pro:proj}
% The contrastive loss can encourage uniformity~\cite{wang2020understanding} and will encourage the distribution of negative samples close to uniform distribution~\cite{ma2023deciphering}. According to Theorem 3, the uniformity of probability distribution will cause entropy to increase, which violates our task I of reducing representation entropy.
% \end{myPro}
% \textit{Proof. }
% We define $\{\Mat{x}_j\}_{j=1}^K$ as the negative samples for instance $\Mat{x}_i$.  
% Then the contrastive loss for negative samples can be defined as:
% \begin{equation}
% \begin{split}
% \label{nega}
% \frac{1}{N}\sum_{i=1}^N-\log \frac{1}{\sum_j\mathrm{exp}({\Mat{x}^\top_i\Mat{x}_j/\tau)}}
%     &=\frac{1}{N}\sum_{i=1}^N\log e^{1/\tau}(\sum_{j=1}^K\mathrm{exp}(-||\Mat{x}_i-\Mat{x}_j||_2^2/2{\tau})\\
%     &=\frac{1}{N}\sum_{i=1}^N\log(\sum_{j=1}^K\mathrm{exp}(-||\Mat{x}_i-\Mat{x}_j||_2^2/2{\tau})+\alpha_0\\
% \end{split}
% \end{equation}
% \hfill $\square$
We consider applying contrastive learning in another latent space $\Mat{z}_G$ that is mapped from $\Mat{h}_G$ by a projection head~\cite{yao2022pcl, chen2020simple}. 
We employ a two-layer $\mathrm{MLP}$ $f_{\theta_p}(\cdot)$ as the projection head:
\begin{gather}
\Mat{z}_G=f_{\theta_p}(\Mat{h}_G). 
\end{gather}
% Then InfoIGL applies contrastive learning and encourages uniformity in new space $\Mat{z}_G$.
% Subsequently, we replace the original embedding $h_G$ with $Z_G$ in our contrastive learning framework for task II. The loss for semantic-wise contrastive learning can be updated as follows:
% \begin{align}
%     \mathcal{L}_{\mathrm{sem}}&=-\frac{1}{N^{tr}}\sum_{i=1}^{N^{tr}}\mathrm{log}\frac{\mathrm{exp}(Z_{G_i}^\top W_c'/ \tau)}{\mathrm{exp}(Z_{G_i}^\top W_c'/\tau)+\sum_{k=1}^{C-1}\mathrm{exp}(Z_{G_i}^\top W_k'/\tau)}
%     \label{equ:loss_sem_}   
% \end{align}

\noindent\textbf{Semantic-level Contrastive Learning. }
\label{sec:semantic_wise}
Semantic-level contrastive learning forces the graph embeddings $z_G$ to be closer to their corresponding category semantics, promoting both inter-class separation and intra-class compactness. {Since the invariance presents a complex nature across different graphs and is difficult to extract, the clustering semantic features for the graphs are more stable and robust.}
For semantic-level contrastive learning, we first introduce the cluster center of each class as the corresponding {semantic}. Formally, we initialize the semantic $\Mat{w}_c$ as the average semantic representation over examples belonging to class c: $\Mat{w}_c=\frac{1}{N_c}\sum_{i=1}^{N_c}(\Mat{z}_{G_i})$, $N_c$ is the number of $\Mat{z}_G$ with label c in a batch. Then we update the current round $\Mat{w}_c^{(r)}$ by calculating the similarity between each instance embedding $\Mat{z}_{G_i}$ and the semantic of last round $\Mat{w}_c^{(r-1)}$:
\begin{equation}
\begin{array}{c}
   m_i^{(r)}=\mathrm{softmax}(\mathrm{cosine}(\Mat{z}_{G_i},\Mat{w}_c^{(r-1)})) \\
   \Mat{w}_c^{(r)}=\sum_{i=1}^{N_c}m_i^{(r)}\cdot \Mat{z}_{G_i}
% \end{split}
\end{array}
\end{equation}
where $\mathrm{cosine}(\cdot)$ denotes the cosine similarity.
Then we define the semantic-level contrastive loss $\mathcal{L}_{\mathrm{sem}}: =$ 
% \begin{equation}
% \begin{split}
%     \mathcal{L}_{\mathrm{sem}}&=-\frac{1}{N^{tr}}\sum_{i=1}^{N^{tr}}\\
% &\mathrm{log}\frac{\mathrm{exp}(\Mat{z}_{G_i}^\top \Mat{w}_c/ \tau)}{\mathrm{exp}(\Mat{z}_{G_i}^\top \Mat{w}_c/\tau)+\sum_{k=1,k\neq c}^{C-1}\mathrm{exp}(\Mat{z}_{G_i}^\top \Mat{w}_k/\tau)}
%     \label{equ:loss_sem}
% \end{split}
% \end{equation}
\begin{equation}
\begin{split}
    -\frac{1}{N^{tr}}\sum_{i=1}^{N^{tr}}\mathrm{log}\frac{\mathrm{exp}(\Mat{z}_{G_i}^\top \Mat{w}_c/ \tau)}{\mathrm{exp}(\Mat{z}_{G_i}^\top \Mat{w}_c/\tau)+\sum_{k=1,k\neq c}^{C-1}\mathrm{exp}(\Mat{z}_{G_i}^\top \Mat{w}_k/\tau)}
    \label{equ:loss_sem}
\end{split}
\end{equation}
where $N^{tr}$ denotes the number of graphs in a batch, $\Mat{w}_c$ denotes the target category semantic of $\Mat{z}_{G_i}$, $C$ denotes the number of classes, $\tau$ is the scale factor. 

\noindent\textbf{Instance-level Contrastive Learning. }
We perform instance-level contrastive learning to fully explore the invariance for prediction. Instance-level contrastive learning aligns the similarity between instances with shared labels, {which provides more detailed and informative representations.} Specifically, we treat instances with shared labels as positive pairs, while viewing those from different classes as negative pairs. The loss function for instance-level contrastive learning can be defined by $\mathcal{L}_{\mathrm{ins}}:=$
\begin{equation}
\begin{aligned}
    -\frac{1}{N^{tr}}\sum_{i=1}^{N^{tr}}
    \mathrm{log}\frac{\mathrm{exp}(\Mat{z}_{G_i}^\top \Mat{z}_{G_+}/\tau')}{\mathrm{exp}(\Mat{z}_{G_i}^\top \Mat{z}_{G_+}/ \tau')+\sum_{k=1}^{K}\mathrm{exp}(\Mat{z}_{G_i}^\top \Mat{z}_{G_k}/\tau')}
    \label{equ:loss_ins}
\end{aligned}
\end{equation}
where the positive sample $\Mat{z}_{G_+}$ are randomly selected from graphs that belong to the same class as $\Mat{z}_{G_i}$, $K$ denotes the number of negative samples for each graph instance. $\tau'$ is the scale factor.

\noindent\textbf{Discussion for Instance-level Contrastive Learning. }
Instance-level contrastive learning prone to suffer from model collapse~\cite{jing2021understanding} (\ie{samples are mapped to the same point}) due to excessive alignment of positive samples. So we apply instance constraint and hard negative mining to prevent getting stuck in trivial solutions ~\cite{yao2022pcl}.

\noindent
\textbf{Instance constraint. }
Enhancing the uniform distribution of graph embeddings $\Mat{z}_G$ can prevent model collapse from excessive alignment. Here we achieve it by leveraging the uniformity of {semantics $\Mat{w}_c$}, which have been ensured by semantic-level contrastive learning. Thus we utilize instance constraint:
\begin{gather}
\Mat{z}_G' = \lambda_c\cdot \Mat{z}_G+ (1-\lambda_c)\cdot \Mat{w}_c
\end{gather}
where $\Mat{w}_c$ refers to the corresponding semantic belonging to the same class as $\Mat{z}_G$.

\noindent
\textbf{Hard negative mining.} 
 Hard negative pair~\cite{xuan2020hard,robinson2020contrastive} can help the network learn a better decision boundary in contrastive learning. 
To identify hard negative samples for instances $\Mat{z}_G'$ belonging to class $c$, we calculate the distance between semantic $\Mat{w}_c$ and samples from other classes within a batch, then we choose the $K$ nearest ones as hard negative samples $\{\Mat{z}_\mathrm{hard_k}'\}_{k=1}^K$.
Then the loss for instance-level contrastive learning can be modified as $\mathcal{L}_{\mathrm{ins}}:=$
\begin{equation}
\begin{split}
    -\frac{1}{N^{tr}}\sum_{i=1}^{N^{tr}}
    \mathrm{log}\frac{\mathrm{exp}(\Mat{z}_{G_i}^{'\top}\Mat{z}_{G_+}'/\tau')}{\mathrm{exp}(\Mat{z}_{G_i}^{'\top} \Mat{z}_{G_+}'/\tau')+\sum_{k=1}^{K}\mathrm{exp}(\Mat{z}_{G_i}^{'\top}\Mat{z}_{hard_k}'/\tau')}
    \label{equ:loss_ins_}
\end{split}    
\end{equation}

\subsubsection{Downstream Task and Overall Framework. }
To align the invariant graph features to the downstream task of graph classification, we define the loss function for prediction as follows:
\begin{gather}
 \mathcal{L}_{\mathrm{pred}} = -\frac{1}{N^{tr} }\sum_{i=1}^{N^{tr}}\Mat{y}_i^\top\mathrm{log}(\theta(\Mat{h}_{G_i}))
    \label{equ:loss_pre}
\end{gather}
where $\Mat{y}_i$ is the label of $G_i$, $\theta$ is the classifier for $\Mat{h}_{G_i}$.

% \subsubsection{Overall Framework}
The overview of our proposed framework is illustrated in Figure~\ref{fig:model}. The final loss of our method can be given by: 
\begin{gather}
\mathcal{L}=\mathcal{L}_{\mathrm{pred}}+\lambda_s\mathcal{L}_{\mathrm{sem}} +\lambda_i \mathcal{L}_{\mathrm{ins}} 
\end{gather}
where the hyperparameters $\lambda_s,\lambda_i$ are scaling weights for each loss, which affect the impact of different modules on the model's results. 
The overall training procedure of the proposed InfoIGL is summarized in Algorithm ~\ref{alg1}.

\begin{algorithm}[!htb]
\caption{InfoIGL Pseudocode. } 
\label{alg1}
\begin{algorithmic}[1]
\SetKwInOut{Input}{Input}
\REQUIRE{Training graph dataset $\mathcal{D}^{tr}$, hyperparameters $ \lambda_s, \lambda_i, \lambda_c$, number of epochs $e$, batch size $b$. } 
\ENSURE{graph encoder $\Phi(\cdot)$ (GNN encoder and Attention mechanism), classifier $\theta(\cdot)$. }
\STATE Initialize graph encoder $\Phi(\cdot)$, Projection head $f_{\theta_p}(\cdot)$, classifier $\theta(\cdot)$.
\FOR {epoch in $1,2,...e$  }
\STATE {Sample data batches $ \mathcal{B}= {\mathcal{D}^1, \mathcal{D}^2 ,..., \mathcal{D}^K}$ with batch size $b$ from $\mathcal{D}^{tr}$.}
% % \STATE sample a batch $B_{tr}$ $\leftarrow$ $\{(G_k,\Mat{y}_k)\}_{k=1}^b$ $\subset$ $\mathcal{D}^{tr} $ 
\FOR {$D^{i}$ $\leftarrow\{(G_k,\Mat{y}_k)\}_{k=1}^b$ $\subset$ $\mathcal{B}, i\in 1...K$ }
% % \STATE {\color{gray} \# Extract invariant graph representation. }
\STATE {\color{gray} \#  Compressing redundancy}
\STATE Calculate $\Mat{h}_{G_k}\leftarrow \Phi(G_k)$ .         
% \STATE $\tilde{\Mat{h}_{G_k}}=f_{\theta_m}(\Mat{h}_{G_k})$.         \hfill{\color{gray} \# Attention mechanism}

\STATE {\color{gray} \#  Multi-level contrastive learning}
\STATE Calculate $\Mat{z}_{G_k}=f_{\theta_p}(\Mat{h}_{G_k})$.        
 % \hfill{\color{gray} \# Projection head}
\STATE Aggregate $\Mat{z}_{G_k}$ to $\Mat{w}_c$.
 % \hfill {\color{gray} \#  Get category semantics}
\STATE Calculate the semantic-level contrastive loss $\mathcal{L}_{\mathrm{sem}}$. 
\STATE Calculate $\Mat{z}_{G_k}'=\lambda_c \Mat{z}_{G_k}+ (1-\lambda_c)\Mat{w}_c$. 
\STATE Obtain hard negative samples $\Mat{z}_{hard}'$.  
\STATE Calculate the instance-level contrastive loss $\mathcal{L}_{\mathrm{ins}}$.  
\STATE {\color{gray} \#  Transfer to downstream tasks}
\STATE Calculate the prediction loss $\mathcal{L}_{\mathrm{pred}}$.
\STATE $\mathcal{L}= \mathcal{L}_{\mathrm{pred}}+\lambda_s\mathcal{L}_{\mathrm{sem}} +\lambda_i \mathcal{L}_{\mathrm{ins}}.  $ 
% \hfill {\color{gray} \#  Total loss}
\ENDFOR
\ENDFOR

\STATE Update all the trainable parameters to minimize $\mathcal{L}$

\end{algorithmic}
\end{algorithm}

\subsection{Time and Space Complexity Analysis}

{Let $N$ be the number of graphs, $n$ be the average node number per graph, $l_G, l_A, l_P$ and $d_G$, $d_A$, $d_P$ be the numbers of layers and the embedding dimensions in the GNN backbone, attention mechanism and projection head, respectively, $C$ be the number of class, and $K$ be the number of hard negative samples per instance. The time complexity of the GNN backbone is $O(N nl_G d_G )$. For the attention mechanism, the time complexity is $O(N nl_A d_A )$. For the projection head, since it turns from node level to graph level, the time complexity is $O(N l_P d_P )$. For semantic-level contrastive learning, the time complexity is $O(NC)$. For instance-level contrastive learning, the time complexity is $O(NK)$. Therefore, the time complexity of the whole model is $O(N (nl_G d_G+nl_A d_A+nl_P d_P )+C+K )$, and the order of magnitude is $O(Nn)$. Similarly, the space complexity is also approximately $O(Nn)$ which is about the same as baselines.}

%% file: 5_experiments.tex
\section{Experiments}

In this section, we conduct extensive experiments on multiple benchmarks to answer the following questions:

\begin{itemize}[leftmargin=*]
\item Q1: How effective is InfoIGL compared to existing methods for OOD generalization in graph classification tasks? 
\item Q2: How do reducing redundancy and maximizing mutual information of graphs work respectively?  
\item Q3: How do the different levels of contrastive learning impact InfoIGL’s performance? 
\item Q4: How sensitive is the model to the hyperparameters and GNN backbones? 
\item Q5: Can this model be extended to tackle the graph OOD problem on node classification tasks?
\end{itemize}
We assess the efficacy of InfoIGL across various out-of-distribution (OOD) graph datasets for graph classification tasks in Q1. To address Q2, we perform an ablation study on the redundancy filter and multi-level contrastive learning. For Q3, we further conduct an ablation study examining the impact of semantic-level and instance-level contrastive learning, respectively. Additionally, we explore the sensitivity of InfoIGL to hyperparameters $\lambda_c$, $\lambda_s$, and $\lambda_i$, as well as to different GNN backbones such as GCN, GIN, and GAT. To validate the scalability of InfoIGL, we extend its application to node classification tasks. Visualization techniques are employed to vividly illustrate the significance of InfoIGL in identifying invariant features.
\subsection{Expermental Settings}
\subsubsection{Datasets and Baselines}
\label{setup}
We conduct experiments on one synthetic (\ie{Motif}) and three real-world (\ie{HIV, Molbbbp, and CMNIST}) datasets designed for graph OOD ~\cite{gui2022good} on graph classification tasks, where Motif and CMNIST are evaluated with ROC-AUC while HIV and Molbbbp are evaluated with ACC, following the setting of GOOD~\cite{gui2022good}. Moreover, to answer Q5 and extend our work on node classification tasks, we conduct experiments on two synthetic datasets (\ie{Cora~\cite{yang2016revisiting} and Amazon-Photo~\cite{rozemberczki2021multi}}). The details of the datasets are as follows. 
\begin{itemize}[leftmargin=*]
\item \textbf{Motif~\cite{gui2022good}} is a synthetic dataset motivated by Spurious-Motif~\cite{ying2019gnnexplainer}, graphs of which are generated by connecting a base graph (wheel, tree, ladder, star, and path) and a motif (house, cycle, and crane). The motifs are invariant for prediction while the base graphs may cause distribution shifts. Here, we use the concept shift of ``base" and the covariate shift of ``size'' to create testing datasets.

\item\textbf{ HIV~\cite{gui2022good} and Molbbbp~\cite{hu2020open} }are small-scale molecular datasets adapted from MoleculeNet~\cite{wu2018moleculenet} in the real world, where atoms serve as nodes and chemical bonds serve as edges. Following the setting of GOOD~\cite{gui2022good}, ``scaffold'' and ``size'' are defined as the environmental features to create distribution shifts. Here, we select concept shifts of ``size'' and ``scaffold'' for testing.
\item\textbf{ CMNIST~\cite{gui2022good}} is a real-world dataset created by applying superpixel techniques to handwritten digits, where ``color'' is defined as the environmental features for distribution shifts. Here we use the covariate shift as the OOD testing dataset. In the covariate shift split, digits are colored with seven different colors, with the first five colors, the sixth color, and the seventh color assigned to the training, validation, and test sets, respectively.  
\item\textbf{Cora~\cite{yang2016revisiting} and Amazon-Photo~\cite{rozemberczki2021multi}} are synthetic datasets for node classification tasks with artificial transformation as distribution shifts, following the setting of EERM~\cite{wu2022handling}. The training, valid, and test datasets are created with covariate shifts, which are split with distinct environment IDs.
\end{itemize}
% \label{sta_datasets}
Statistics of these datasets are presented in Table~\ref{table:dataset}. 
\begin{table*}[!htb]
\centering
\renewcommand\arraystretch{1.2}
% \vspace{-2mm}
~\\
\caption{Statistics of multiple graph OOD datasets: Motif, HIV, Molbbbp, and CMNIST for graph classification tasks, and Cora and Amazon-Photo for node classification tasks. ``Train'', ``Val'', and ``Test'' denote the numbers of graphs in the training set, OOD validation set, and OOD test set, respectively. ``Classes'' and ``Metrics'' refer to the number of classes and evaluation metrics used for these datasets. ``Shift'' indicates the type of distribution shifts.}
~\\
\vspace{2mm}
% \vspace{-2mm}
\label{table:dataset}
\setlength{\tabcolsep}{5mm}{\begin{tabular}{lcccccc}
\toprule
\textbf{Dataset} & \#\textbf{Train} & \#\textbf{Val} & \#\textbf{Test} & \#\textbf{Classes}& \#\textbf{Metrics} &\#\textbf{Shift} \\
\toprule
Motif-size& 18000 & 3000 & 3000 & 3 & ACC&covariate \\
Motif-base& 12600 & 6000 & 6000 & 3& ACC&concept \\
\hline
HIV-size& 14454 & 9956 & 10525 & 2 &ROC-AUC&concept\\
HIV-scaffold& 15209 & 9365 & 10037 & 2 &ROC-AUC&concept\\
\hline

Molbbbp-size & 1631 & 204 & 204 & 2 &ROC-AUC&concept\\
Molbbbp-scaffold & 1631 & 204 & 204 & 2&  ROC-AUC&concept\\
\hline
CMNIST-color & 42000 & 14000 & 14000 & 10 &ACC&covariate\\
\hline
Cora & - & - & - & 10 &ACC&covariate\\
Amazon-Photo&-&-&-&10&ACC&covariate\\
\bottomrule
\end{tabular}}
\end{table*}

We compare our InfoIGL against diverse graph OOD generalization baselines: \textbf{Optimization methods:} ERM, IRM~\cite{arjovsky2019invariant}, VREx ~\cite{krueger2021out}, GroupDRO~\cite{sagawa2019distributionally}, FLAG~\cite{kong2022robust}; \textbf{Causal learning:} DIR~\cite{DIR}, CAL~\cite{sui2022causal}, GREA~\cite{liu2022graph}, CIGA~\cite{chenlearning}, Disc~\cite{fan2022debiasing}; \textbf{Stable learning:} StableGNN~\cite{fan2021generalizing},  OOD-GNN~\cite{li2021ood}; and \textbf{Data manipulation method:} GSAT~\cite{miao2022interpretable},
DropEdge~\cite{rong2019dropedge}, M-Mixup~\cite{wang2021mixup}, G-Mixup~\cite{han2022g}. Since the practical implementation of InfoIGL involves graph contrastive learning to maximize mutual information, we also adopt \textbf{classical graph contrastive learning} methods as benchmarks, including CNC~\cite{zhang2022correct}, GMI~\cite{peng2020graph}, Infograph~\cite{sun2019infograph}, GraphCL~\cite{hafidi2007graphcl}. Optimization methods aim to design optimization objectives to enhance the robustness of the model across different environments. Causal learning utilizes the causal theory to extract causal features that play a key role in model prediction and ignore non-causal features. 
Stable learning is committed to independently extracting stable features across different environments through sample reweighting,
Data manipulation is dedicated to generating diverse augmented data to increase the diversity of data distribution. While
classical graph contrastive learning methods incorporate self-supervised learning and GNN to enhance the quality of graph representations.
\subsubsection{Implementation Details}

Our code is implemented based on PyTorch Geometric. For all the experiments, we use the Adam optimizer, where the initial and minimum learning rate are searched within $\{0.01, 0.001, 0.0001\}$ and $\{0.001, 0.00001, 0.000001\}$, respectively. We select embedding dimensions from $\{32, 64, 128, 300\}$ and choose batch sizes from $\{64, 128, 256, 512, 1024\}$. The dropout ratio is searched within $\{0.1, 0.3, 0.5\}$ while $\lambda_c, \lambda_s,\lambda_i$ are searched within $\{0.1,0.2, ... ,0.9\}$. We adopt grid search to tune the hyperparameters and list the details of hyperparameters for InfoIGL in Table~\ref{table:hyper_all}. 
\begin{table}[!htb]
\centering
\caption{The hyperparameters for InfoIGL on different datasets, where ``sca'' denotes ``scaffold''. }
\vspace{2mm}
\renewcommand\arraystretch{1.2}
\label{table:hyper_all}
\setlength{\tabcolsep}{1mm}{\begin{tabular}{lccccccc}
\toprule
\multirow{2}*{HP} & \multicolumn{2}{c}{Motif}  & \multicolumn{2}{c}{HIV} & \multicolumn{2}{c}{Molbbbp} & CMNIST  \\
\cmidrule(lr){2-3}\cmidrule(lr){4-5}\cmidrule(lr){6-7}\cmidrule(lr){8-8}
& size & base & size & sca & size & sca & color\\ 
\toprule
layers & 
3 & 
3 & 
3 & 
3 & 
2&
2&
5
 \\
emb-dim  & 
64 &
128 & 
300 & 
128 & 
128&
300&
32
 \\
max-epoch   & 
200 & 
200 & 
100 & 
200& 
100&
100&
150
 \\
% pretrain &
% 40&
% 40&
% 80&
% 40&
% 20&
% 80&
% 60

% \\
batch size  & 
1024 & 
128 & 
256 & 
1024 & 
64 &
1024&
256
 \\
ini-lr& 
1e-3 &
1e-3 & 
1e-3 & 
1e-2 & 
1e-2 &
1e-4 &
1e-3 
 \\
min-lr& 
1e-3 &
1e-6 & 
1e-6 & 
1e-6 & 
1e-6 &
1e-6 &
1e-3
 \\

decay & 
0 &
1e-1 &
1e-2 & 
1e-5 & 
1e-5 &
0 &
0
 \\

% drop ratio & 
% 0.5 &
% 0.5 &
% 0.3 & 
% 0.3 & 
% 0.3 &
% 0.3 &
% 0.5\\

$\lambda_c$& 
0.7 &
0.7 & 
0.7 & 
0.7 & 
0.2&
0.7&
0.7\\

$\lambda_s$& 
0.8 &
0.5 & 
0.5 & 
0.5 & 
0.2&
0.2&
0.5
 \\
 $\lambda_i$& 
0.2 &
0.5 & 
0.5 & 
0.1 & 
0.2&
0.2&
0.1
 \\
 
\bottomrule
\end{tabular}}
\end{table}

\subsection{Overall Results (Q1)}

\begin{table*}[!htb]
\renewcommand\arraystretch{1.2}
% \vspace{-10pt}
\caption{Performance of different methods on synthetic (Motif) and real-world (HIV, Molbbbp, CMNIST) datasets. The best results are in \textbf{bold}, and the runner-up results are \underline{underlined}.}
\vspace{2mm}
\label{table:main_all}
\centering
\setlength{\tabcolsep}{3mm}{\begin{tabular}{lccccccc}
\toprule
\multirow{2}*{methods} & \multicolumn{2}{c}{Motif} & \multicolumn{2}{c}{HIV}  & \multicolumn{2}{c}{Molbbbp} & CMNIST  \\
\cmidrule(lr){2-3}\cmidrule(lr){4-5}\cmidrule(lr){6-7}\cmidrule(lr){8-8}
& size & base & size & scaffold & size & scaffold & color\\ 
% \cline{2-3}

\toprule

ERM   & 70.75\tiny{$\pm$0.56} & 81.44\tiny{$\pm$0.45}& 63.26\tiny{$\pm$2.47} & 72.33\tiny{$\pm$1.04}  & 
78.29\tiny{$\pm$3.76}&
68.10\tiny{$\pm$1.68} &28.60\tiny{$\pm$1.87}
 \\
IRM & 69.77\tiny{$\pm$0.88} & 80.71\tiny{$\pm$0.46}  & 59.90\tiny{$\pm$3.15} & 72.59\tiny{$\pm$0.45}  &
77.56\tiny{$\pm$2.48}&
67.22\tiny{$\pm$1.15} &
27.83\tiny{$\pm$2.13}
 \\
GroupDRO & 69.98\tiny{$\pm$0.86} & 81.43\tiny{$\pm$0.70} &61.37\tiny{$\pm$2.79} & \textbf{73.64\tiny{$\pm$0.86}} &  
79.27\tiny{$\pm$2.43}&
66.47\tiny{$\pm$2.39} &
29.07\tiny{$\pm$3.14}
 \\
VREx & 70.24\tiny{$\pm$0.72} & 81.56\tiny{$\pm$0.35} &60.23\tiny{$\pm$1.70} & 72.60\tiny{$\pm$0.82} & 78.76\tiny{$\pm$2.37}&
68.74\tiny{$\pm$1.03} &
28.48\tiny{$\pm$2.87}
 \\

 FLAG  &56.26\tiny{$\pm$3.98}&
72.29\tiny{$\pm$1.31}& 
66.44\tiny{$\pm$2.32}& 70.45\tiny{$\pm$1.55} & 
79.26\tiny{$\pm$2.26}&
67.69\tiny{$\pm$2.36} &32.30\tiny{$\pm$2.69}
 \\

\toprule

DIR & 54.96\tiny{$\pm$9.32} & 82.96\tiny{$\pm$4.47} & 72.61\tiny{$\pm$2.03} & 69.05\tiny{$\pm$0.92}  &
76.40\tiny{$\pm$4.43} &
66.86\tiny{$\pm$2.25} &
33.20\tiny{$\pm$6.17}
 \\

CAL &66.64\tiny{$\pm$2.74} & 81.94
\tiny{$\pm$1.20} &\underline{83.33\tiny{$\pm$2.84}}&
73.05\tiny{$\pm$1.86} & 79.20
\tiny{$\pm$3.81} &
67.37\tiny{$\pm$3.61}&
27.99{\tiny{$\pm$3.24}}\\

GREA & \underline{73.31\tiny{$\pm$1.85}} &
80.60\tiny{$\pm$2.49} 
& 66.48\tiny{$\pm$4.13} & 70.96\tiny{$\pm$3.16} & 
77.34\tiny{$\pm$3.52}&
\underline{69.72\tiny{$\pm$1.66}}&
29.02\tiny{$\pm$3.26}
 \\
CIGA  & 70.65\tiny{$\pm$4.81} &
75.01\tiny{$\pm$3.56} & 
65.98\tiny{$\pm$3.31} & 
64.92\tiny{$\pm$2.09} & 
76.08\tiny{$\pm$1.21} &
66.43\tiny{$\pm$1.99} &
23.36\tiny{$\pm$9.32}
 \\

DisC & 53.34\tiny{$\pm$13.71} &
76.70\tiny{$\pm$0.47}
& 56.59\tiny{$\pm$10.09} & 67.12\tiny{$\pm$2.11}  & 
75.68\tiny{$\pm$3.16}&
60.72\tiny{$\pm$0.89} &24.99\tiny{$\pm$1.78}
 \\

\toprule
GSAT & 64.16\tiny{$\pm$3.35}&
\underline{83.71\tiny{$\pm$2.30}} & 65.63\tiny{$\pm$0.88} & 68.88\tiny{$\pm$1.96}  &
75.63\tiny{$\pm$3.83}&
66.78\tiny{$\pm$1.45} &28.17\tiny{$\pm$1.26}
 \\
 
DropEdge  & 55.27\tiny{$\pm$5.93}&
70.84\tiny{$\pm$6.81} & 
54.92\tiny{$\pm$1.73} & 66.78\tiny{$\pm$2.68} &
78.32\tiny{$\pm$3.44}&
66.49\tiny{$\pm$1.55} &
22.65\tiny{$\pm$2.90}
 \\
 
M-Mixup & 67.81\tiny{$\pm$1.13} & 77.63\tiny{$\pm$0.57} & 64.87\tiny{$\pm$1.77} & 72.03\tiny{$\pm$0.53}  & 
78.92\tiny{$\pm$2.43}&
68.75\tiny{$\pm$1.03} &
26.47\tiny{$\pm$3.45}
 \\

G-Mixup & 59.92\tiny{$\pm$2.10} & 74.66\tiny{$\pm$1.89}  & 70.53\tiny{$\pm$2.02} & 71.69\tiny{$\pm$1.74} &78.55\tiny{$\pm$4.16} & 67.44\tiny{$\pm$1.62} & 31.85\tiny{$\pm$5.82}
 \\
\toprule
 
OOD-GNN  & 68.62\tiny{$\pm$2.98} & 74.62\tiny{$\pm$2.66} & 57.49\tiny{$\pm$1.08} & 70.45\tiny{$\pm$2.02} & 
79.48\tiny{$\pm$4.19}&
66.72\tiny{$\pm$1.23} &
26.49\tiny{$\pm$2.94}
 \\

StableGNN   & 59.83\tiny{$\pm$3.40} &
73.04\tiny{$\pm$2.78}  & 58.33\tiny{$\pm$4.69} & 68.23\tiny{$\pm$2.44}&
77.47\tiny{$\pm$4.69}&
66.74\tiny{$\pm$1.30} &28.38\tiny{$\pm$3.49}
 \\
%  CAL~\cite{sui2022causal}   & 62.36\tiny{$\pm$1.42} & 72.61\tiny{$\pm$1.84} & 66.64\tiny{$\pm$2.74} &68.54\tiny{$\pm$2.14} & 
% \underline{79.50\tiny{$\pm$4.81}}&
% 68.06\tiny{$\pm$2.60} &42.48\tiny{$\pm$0.48}
\toprule
CNC  & 66.52\tiny{$\pm$3.12} &
82.51\tiny{$\pm$1.26}  & 70.68\tiny{$\pm$2.15} & 66.53\tiny{$\pm$2.19}&
76.19\tiny{$\pm$3.52}&
68.16\tiny{$\pm$1.25} &32.41\tiny{$\pm$1.28}
 \\
GMI  & 67.90\tiny{$\pm$1.46} &
79.52\tiny{$\pm$0.45}  &74.34\tiny{$\pm$0.55} & \underline{73.44\tiny{$\pm$0.35}}&
77.67\tiny{$\pm$0.30}&
69.38\tiny{$\pm$1.02} &30.24\tiny{$\pm$5.98}
 \\
InfoGraph  & 67.49\tiny{$\pm$2.54} &
75.57\tiny{$\pm$0.88}  & 74.63\tiny{$\pm$0.80} & 71.41\tiny{$\pm$0.82}&
\underline{80.82\tiny{$\pm$0.49}}&
70.39\tiny{$\pm$1.34} &\underline{33.84\tiny{$\pm$1.52}}
 \\

GraphCL  & 66.90\tiny{$\pm$2.80} &
74.40\tiny{$\pm$0.90}& 
77.13\tiny{$\pm$0.17}&
72.94\tiny{$\pm$0.68}&
80.64\tiny{$\pm$0.78}&
69.36\tiny{$\pm$1.32} &32.81\tiny{$\pm$1.71}
 \\

\rowcolor{mygray}
InfoIGL(ours) &\textbf{85.53\tiny{$\pm$2.37}} & \textbf{92.51
\tiny{$\pm$0.16}}&\textbf{93.15\tiny{$\pm$0.77}} 
 & 72.37\tiny{$\pm$1.63} & \textbf{83.39
\tiny{$\pm$2.76}} &
\textbf{77.05
\tiny{$\pm$2.24}} &
\textbf{38.93{\tiny{$\pm$1.11}}}\\

\hline
\rowcolor{mygray}improvemrnt&$ \uparrow 12.22 \%$ & $\uparrow 8.80 \%$  & $ \uparrow 9.82 \%$
 & $\downarrow 1.07 \%$ &  $\uparrow 2.57 \%$ & $\uparrow 7.33\%$&$\uparrow 5.09 \%$
 \\ 
\bottomrule
\end{tabular}}
\end{table*}

We train and evaluate our proposed InfoIGL, together with all the baselines, 10 times to obtain the average performance (mean ± standard deviation). 
% The overall performance of InfoIGL and baselines for tackling graph OOD problem is reported in Table~\ref{table:main_all}, where the best and second best results are highlighted with bold and underlined numbers respectively. The last row is the improvement of our approach compared to the best baseline. 
It can be observed from Table~\ref{table:main_all} that optimization methods exhibit stable performance with moderate accuracy and low variance, while causal learning baselines show unstable performance with undulating accuracy and high variance. Besides, stable learning and data manipulation baselines perform relatively poorly compared to other baselines. Additionally, traditional graph contrastive learning methods can partially combat distributional shifts, but their effectiveness is not as strong as InfoIGL since they were not designed specially to extract invariant features. These observations indicate that almost all of the baselines have their limitations for graph OOD generalization. Our proposed framework, InfoIGL, achieves state-of-the-art performance on diverse datasets with low variance, outperforming the strongest baseline by 9.82\% on HIV (size) and 12.22\% on Motif (size). We conduct multiple tests with p-value$<0.05$,
demonstrating that the performance improvements of InfoIGL are statistically significant. The relatively weak performance of InfoIGL on HIV (scaffold) can be attributed to the Variability and complexity of molecular structures. These results demonstrate the effectiveness of InfoIGL in extracting stable and invariant graph representations on both concept and covariate shifts for graph classification tasks.

\subsection{Ablation Study for Q2}

\begin{table*}[!htb]
\renewcommand\arraystretch{1.2}
\centering
\caption{Results of ablation experiments on redundancy filter and contrastive learning. InfoIGL-R is the variant of InfoIGL with merely a redundancy filter, while InfoIGL-C is that with contrastive learning only. The best results are in \textbf{bold}. The results of methods that are superior to that of ERM are marked with $\uparrow$. }
\vspace{2mm}
\label{table:Ablation_2}
\setlength{\tabcolsep}{3mm}{\begin{tabular}{lccccccc}
\toprule
\multirow{2}*{methods} & \multicolumn{2}{c}{Motif}  & \multicolumn{2}{c}{HIV} & \multicolumn{2}{c}{Molbbbp} & CMNIST  \\
\cmidrule(lr){2-3}\cmidrule(lr){4-5}\cmidrule(lr){6-7}\cmidrule(lr){8-8}
& size & base & size & scaffold & size & scaffold & color\\ 
% \cline{2-3}
\toprule

ERM
& 70.75\tiny{$\pm$0.56} & 81.44\tiny{$\pm$0.45}& 63.26\tiny{$\pm$2.47} & 72.33\tiny{$\pm$1.04}  & 
78.29\tiny{$\pm$3.76}&
68.10\tiny{$\pm$1.68} &28.60\tiny{$\pm$1.87}
 \\
InfoIGL-R  & 
{69.69\tiny{$\pm$6.24}} $\downarrow$ & 
{87.14\tiny{$\pm$0.88}} $\uparrow$&
{76.99\tiny{$\pm$2.55}}$\uparrow$ & 
{71.56\tiny{$\pm$1.96}} $\downarrow$& 
{79.72\tiny{$\pm$3.50}} $\uparrow$&
{74.48\tiny{$\pm$1.00}}$\uparrow$&
{34.54\tiny{$\pm$2.11}}$\uparrow$
 \\

InfoIGL-C  & 
{68.01\tiny{$\pm$2.09}}$\downarrow$ & 
{86.63\tiny{$\pm$1.33}}$\uparrow$& 
{72.81\tiny{$\pm$2.92}}$\uparrow$ & 
{68.02\tiny{$\pm$2.28}}$\downarrow$& 
{75.32\tiny{$\pm$1.38}}$\downarrow$& 
{65.62\tiny{$\pm$1.07}}$\downarrow$& 
{33.40\tiny{$\pm$2.10}}$\uparrow$
 \\
 % \hline
InfoIGL & 
\textbf{85.53\tiny{$\pm$2.37}}$\uparrow$ & 
\textbf{92.51\tiny{$\pm$0.16}}$\uparrow$& 
\textbf{93.15\tiny{$\pm$0.77}} $\uparrow$&
\textbf{72.37\tiny{$\pm$1.63}}$\downarrow$& 
\textbf{83.39\tiny{$\pm$2.76}} $\uparrow$& 
\textbf{77.05\tiny{$\pm$2.24}}$\uparrow$& 
\textbf{38.93\tiny{$\pm$1.11}}$\uparrow$
 \\
 \hline
% improvemrnt & $ \uparrow 15.84 \%$ & $\uparrow 5.37 \%$ &$ \uparrow 16.16 \%$
%  & $\uparrow 0.81 \%$ &  $\uparrow 3.67 \%$ & $\uparrow 2.57 \%$&$\uparrow 4.39 \%$
%  \\ 
% \bottomrule
\end{tabular}}
\end{table*}
    To validate the significance of each task individually, we conduct separate ablation studies on the redundancy filter and contrastive learning. Specifically, we compare InfoIGL with two variants: (1)  InfoIGL-R: which includes only the redundancy filter with attention mechanism, and (2) InfoIGL-C, which focuses solely on contrastive learning. As is shown in Table ~\ref{table:Ablation_2}, InfoIGL-R outperforms ERM on most of the datasets, demonstrating the effectiveness of reducing redundancy. However, its performance falls short of ERM on Motif-size and HIV-scaffold, which means that the variance can not be identified accurately by compressing redundancy merely, underscoring the significance of optimization from contrastive learning. In contrast, InfoIGL-C yields poorer results than ERM on several datasets, such as Molbbbp and motif-size, shedding light on the negative impact of spurious features and the significance of compressing redundancy based on information bottleneck theory.

\subsection{Ablation Study for Q3}
\begin{table*}[!htb]
\renewcommand\arraystretch{1.2}
\centering
\caption{Results of ablation experiments on semantic-level and instance-level contrastive learning. InfoIGL-S is the variant of InfoIGL with contrastive learning merely from the semantic level, while InfoIGL-I is that with contrastive learning from the instance level only. The best results are in \textbf{bold}. The results of methods that are superior to that of ERM are marked with $\uparrow$.}
\vspace{2mm}
\label{table:Ablation}
\setlength{\tabcolsep}{3mm}{\begin{tabular}{lccccccc}
\toprule
\multirow{2}*{methods} & \multicolumn{2}{c}{Motif}  & \multicolumn{2}{c}{HIV} & \multicolumn{2}{c}{Molbbbp} & CMNIST  \\
\cmidrule(lr){2-3}\cmidrule(lr){4-5}\cmidrule(lr){6-7}\cmidrule(lr){8-8}
& size & base & size & scaffold & size & scaffold & color\\ 
% \cline{2-3}
\toprule

InfoIGL-N  & 
69.69\tiny{$\pm$6.24} & 
87.14\tiny{$\pm$0.88} &
76.99\tiny{$\pm$2.55} & 
71.56\tiny{$\pm$1.96} & 
79.72\tiny{$\pm$3.50} &
74.48\tiny{$\pm$1.00} &
34.54\tiny{$\pm$2.11}
 \\
InfoIGL-S & 
{84.77\tiny{$\pm$2.10} }$\uparrow$ & 
{89.93\tiny{$\pm$0.93}} $\uparrow$ & 
{87.30\tiny{$\pm$1.21}} $\uparrow$& 
{72.12\tiny{$\pm$1.87} }$\uparrow$ & 
{81.97\tiny{$\pm$1.79} }$\uparrow$ & 
{76.76\tiny{$\pm$3.66} }$\uparrow$ & 
{37.31\tiny{$\pm$1.50} }$\uparrow$
 \\
InfoIGL-I  & 
{80.05\tiny{$\pm$2.99}}$\uparrow$ & 
{90.36\tiny{$\pm$1.54}}$\uparrow$ & 
{91.38\tiny{$\pm$2.38}}$\uparrow$ & 
{63.70\tiny{$\pm$5.45}}$\downarrow$ & 
{74.91\tiny{$\pm$2.07}}$\downarrow$ & 
{70.11\tiny{$\pm$2.02}}$\downarrow$ & 
{35.67\tiny{$\pm$1.19}}$\uparrow$
 \\
 % \hline
InfoIGL & 
\textbf{85.53\tiny{$\pm$2.37}}$\uparrow$ & 
\textbf{92.51\tiny{$\pm$0.16}}$\uparrow$ & 
\textbf{93.15\tiny{$\pm$0.77}}$\uparrow$ &
\textbf{72.37\tiny{$\pm$1.63}}$\uparrow$ & 
\textbf{83.39\tiny{$\pm$2.76}}$\uparrow$ & 
\textbf{77.05\tiny{$\pm$2.24}}$\uparrow$ & 
\textbf{38.93\tiny{$\pm$1.11}}$\uparrow$ 
 \\
 \hline
% improvemrnt & $ \uparrow 15.84 \%$ & $\uparrow 5.37 \%$ &$ \uparrow 16.16 \%$
%  & $\uparrow 0.81 \%$ &  $\uparrow 3.67 \%$ & $\uparrow 2.57 \%$&$\uparrow 4.39 \%$
%  \\ 
% \bottomrule
\end{tabular}}
\end{table*}
\begin{figure*}%靠文字内容的右侧
% \vspace{-10pt}
\centering
\includegraphics[width=1\linewidth]{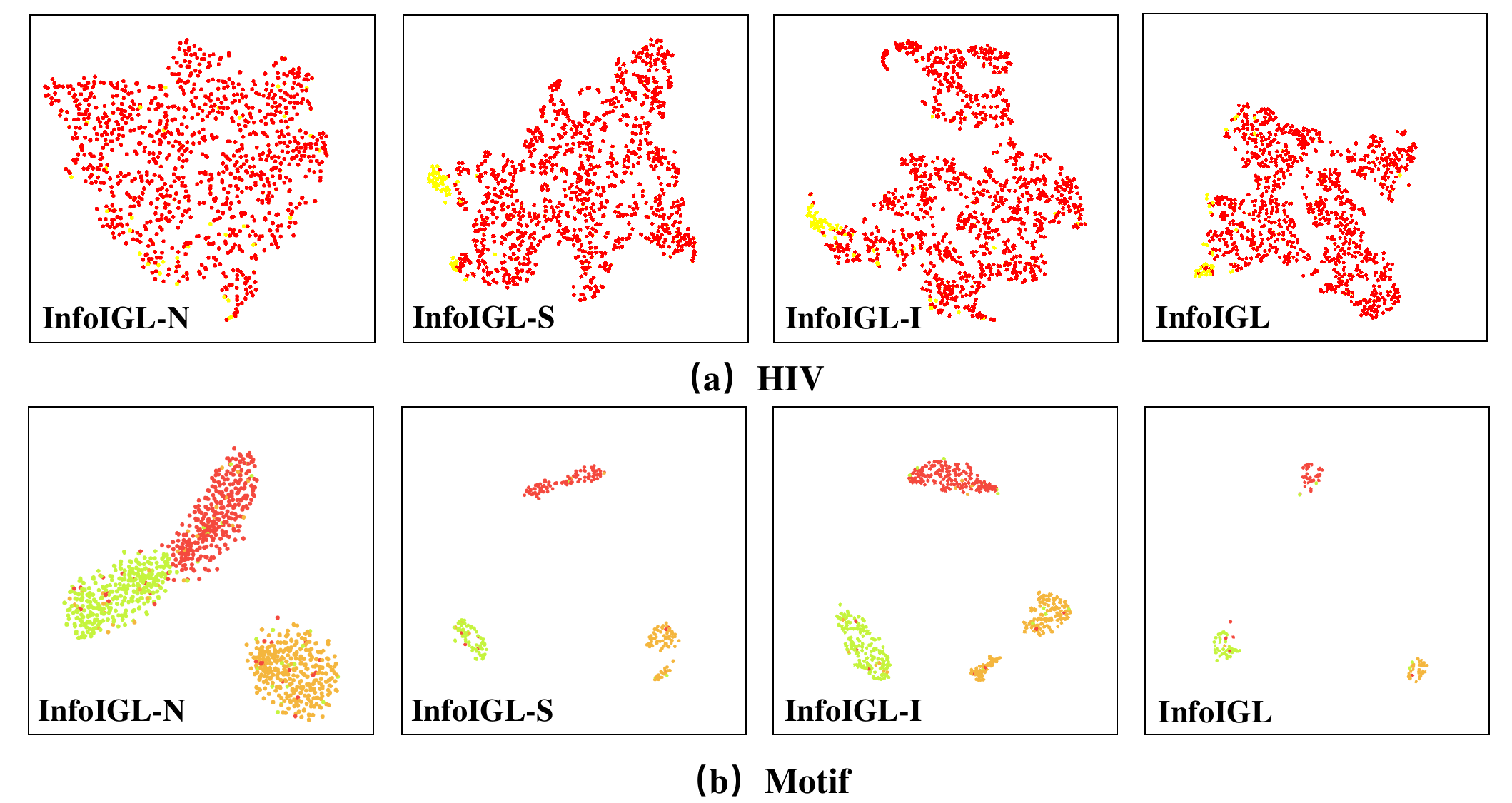}
% \vspace{-10pt}
\caption{The t-SNE visualizations for different levels of contrastive learning.}
\label{fig:visualization}
% \vspace{-10pt}
\end{figure*}
 We perform an ablation study to analyze the impacts of semantic-level and instance-level contrastive learning respectively. Specifically, we compare InfoIGL with three variants: (1) InfoIGL-N, which does not use contrastive learning; (2) InfoIGL-S, which employs semantic-level contrastive learning only; and (3) InfoIGL-I, which applies instance-level contrastive learning only. We report the results in Table~\ref{table:Ablation}. Our observations are as follows: 1) Merely applying semantic-level contrastive learning causes performance degradation on the Molbbbp dataset, which confirms the importance of extracting invariance on local features by instance-level contrastive learning. 
2) Applying instance-level contrastive learning solely falls short of ERM on the datasets of Molbbbp and HIV-scaffold, highlighting its tendency to overlook global features of categories for prediction. 
3) InfoIGL with both semantic and instance-level contrastive learning outperforms all of the three variants across diverse datasets, proving that jointly optimizing the two contrastive losses can inspire their individual potentials. The Semantic- and instance-level contrastive learning can promote each other and extract invariance for prediction to the greatest extent.

 Additionally, we employ the t-SNE technique to visualize embeddings of graph instances on HIV~\cite{gui2022good} and Motif~\cite{gui2022good} datasets (\cf Figure~\ref{fig:visualization}), where the four variants (InfoIGL-N, InfoIGL-S, InfoIGL-I, and InfoIGL) are compared. The results reveal that compared to InfoIGL-N, the embeddings obtained by InfoIGL-S and InfoIGL-I exhibit a more compact clustering pattern, reflecting the efficacy of semantic- and instance-level contrastive learning in aligning shared information, respectively. Furthermore, InfoIGL exhibits the best convergence effect, as its embeddings are more tightly clustered than those produced by the other variants. By promoting inter-class separation and intra-class compactness, the embeddings of graphs {share a greater amount of information}, indicating that InfoIGL can extract improved invariance within a class. This observation highlights the potential of maximizing predictive information of invariance by maximizing the alignment of samples from the same class.

\begin{figure*}[htb]
% \vspace{-6mm}
    \centering
    \includegraphics[width=1\linewidth]{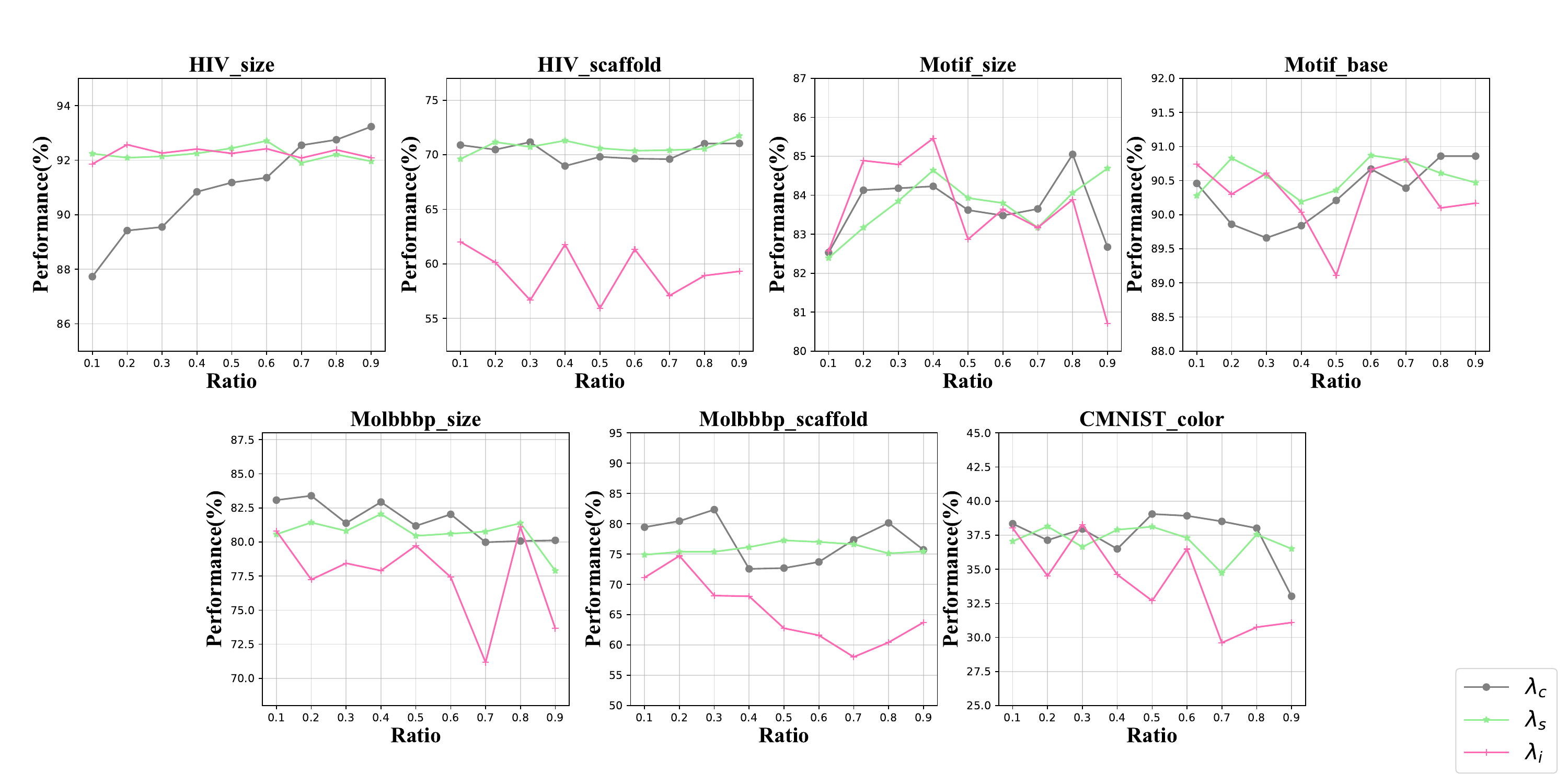}
    \vspace{-4mm}
    \caption{Sensitivity analysis of hyperparameters $\lambda_c, \lambda_s, \lambda_i$}
    \label{fig:sensi}
    \vspace{-4mm}
\end{figure*}

\subsection{Sensitive Analysis (Q4)}
To assess the sensitivity of InfoIGL to its hyperparameters, namely $\lambda_c$ for instance constraint and $\lambda_s$ and $\lambda_i$ for contrastive loss respectively, we conduct sensitivity analysis experiments by tuning these hyperparameters within the range of $\{0.1, 0.2, ..., 0.9\} $ under the controlled experimental setting. Specifically, when adjusting a specific hyperparameter, we fix the remaining hyperparameters at the values that yield the best performance. The results are presented in Figure~\ref{fig:sensi}. It is noteworthy that InfoIGL is relatively insensitive to $\lambda_c$ {(for instance constraint)} and $\lambda_s$ {(for semantic-level contrastive learning)} in gray and green curves, as adjusting their values does not cause significant fluctuations in InfoIGL's performance. While $\lambda_i$ ({for instance-level contrastive learning}) has a significant impact on the InfoIGL and requires fine-tuning.  {
We analyze the reasons behind these phenomena carefully.
\begin{itemize}[leftmargin=*]
\item The significant sensitivity of $\lambda_i$ demonstrates that the instance-level contrastive learning is fragile when applied to complex graph instances. The hyperparameter $\lambda_i$ needs careful finetuning to accurately identify the invariance of the graphs, due to the intricate nature of graph data.
\item The minor influence of $\lambda_s$ demonstrates that the semantic-level contrastive learning is stable and not sensitive to this hyperparameter. This method effectively learns semantic-level invariant features, benefiting from the robustness of global category semantics.
\item The relative smooth curve for $\lambda_c$ confirms that constraining instance-level representations with robust category semantics can facilitate smoother training. Introducing stable category information into complex data instances can enhance prediction accuracy.
    \end{itemize}}

Furthermore, the results demonstrate the negative impact of excessively large values for $\lambda_i$ in InfoIGL. For instance, the performance of InfoIGL on Motif (size) and CMNIST (color) drops significantly when $\lambda_i$ exceeds 0.6. By comparing the performance of InfoIGL across different datasets under different hyperparameter settings, we can identify the optimal hyperparameters for each dataset. For example, $\lambda_s$ ranging from 0.3 to 0.7 and $\lambda_c$ ranging from 0.5 to 0.8 are more suitable hyperparameter values.

We also conduct experiments to evaluate how sensitive is InfoIGL to the choice of graph neural network architectures (GCN, GIN, and GAT). The results are listed in Table~\ref{table:Ablation_GNN}. As shown in Table~\ref{table:Ablation_GNN}, InfoIGL-GCN, InfoIGL-GIN, and InfoIGL-GAT are competent on Motif and CMNIST datasets while they far surpass the baseline ERM. The results demonstrate the effectiveness of our method, irrespective of the choice of GNN backbones.

\begin{table}[!htb]
\renewcommand\arraystretch{1.2}
\centering
\vspace{2mm}
\caption{Results of experiments with different backbones, including GCN, GIN, and GAT. }
\vspace{3mm}
\label{table:Ablation_GNN}
\setlength{\tabcolsep}{3mm}{\begin{tabular}{lccc}
\toprule
\multirow{2}*{methods} & \multicolumn{2}{c}{Motif}  & CMNIST  \\
\cmidrule(lr){2-3}\cmidrule(lr){4-4}
& size & base & color\\ 
% \cline{2-3}
\toprule

ERM
& 70.75\tiny{$\pm$0.56} & 81.44\tiny{$\pm$0.45} &28.60\tiny{$\pm$1.87}
 \\
InfoIGL-GCN  & 
86.53\tiny{$\pm$2.15} & 
91.56\tiny{$\pm$0.91} &
38.30\tiny{$\pm$0.76}
 \\

InfoIGL-GIN  & 
85.53\tiny{$\pm$2.37} & 
92.51\tiny{$\pm$0.16}& 
38.93\tiny{$\pm$1.11}
 \\
 % \hline
InfoIGL-GAT & 
84.66\tiny{$\pm$1.23}& 
90.32\tiny{$\pm$1.45} & 
37.51\tiny{$\pm$2.07}
 \\
 \hline
% improvemrnt & $ \uparrow 15.84 \%$ & $\uparrow 5.37 \%$ &$ \uparrow 16.16 \%$
%  & $\uparrow 0.81 \%$ &  $\uparrow 3.67 \%$ & $\uparrow 2.57 \%$&$\uparrow 4.39 \%$
%  \\ 
% \bottomrule
\end{tabular}}
\end{table}

\begin{figure*}[htb]
% \vspace{-6mm}
    \centering
    \includegraphics[width=0.9\linewidth]{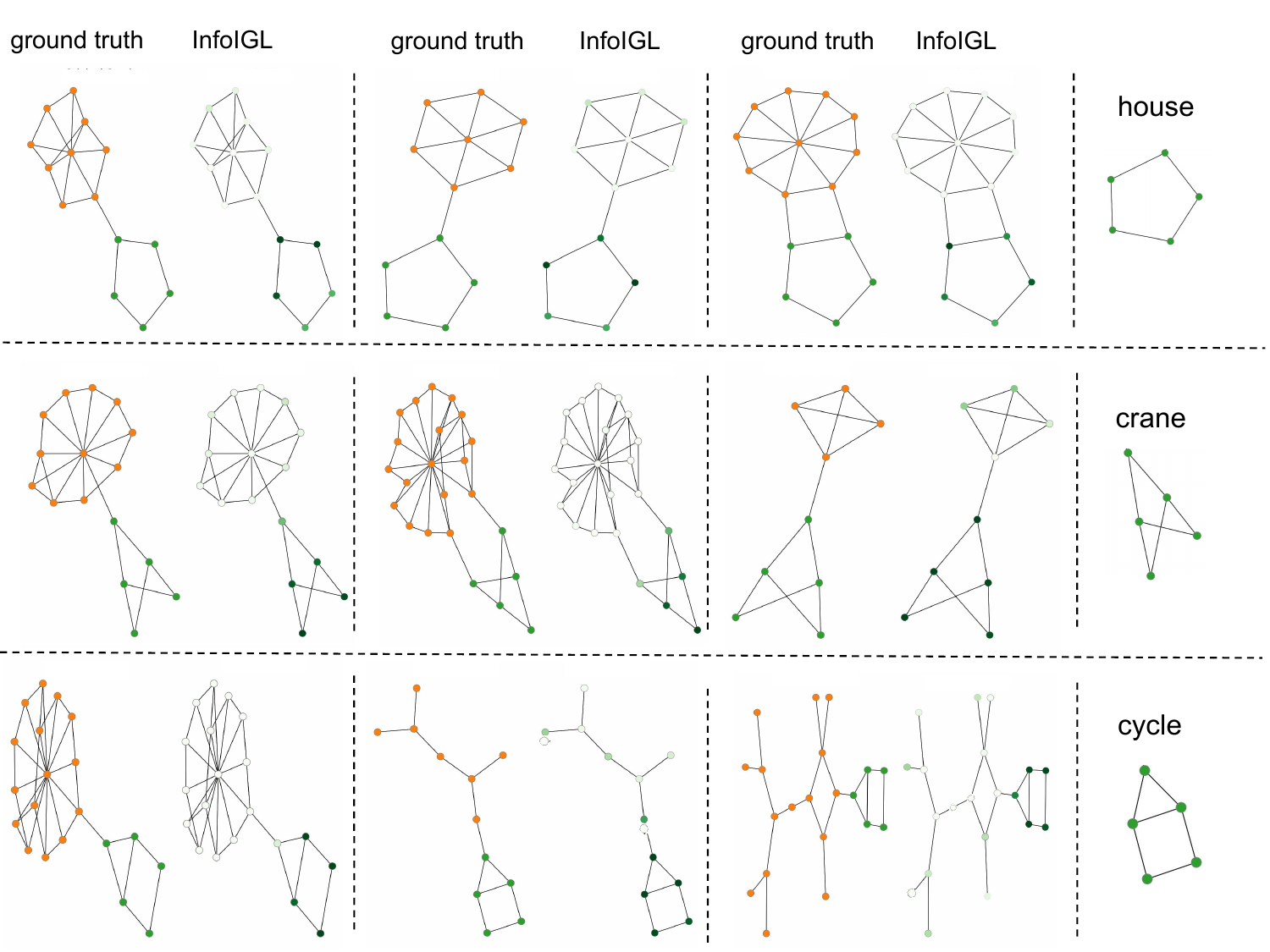}
    \vspace{2mm}
    \caption{The invariance obtained by InfoIGL.}
    \label{fig:visual}
    % \vspace{-4mm}
\end{figure*}

\begin{table*}[htb]
\renewcommand\arraystretch{1.2}
\centering
\vspace{2mm}
\caption{Results of experiments of InfoIGL on node classification tasks on Cora and Amazon-Photo datasets. The best results are in \textbf{bold}.  }
\vspace{2mm}
\label{table:node}
\setlength{\tabcolsep}{4mm}{\begin{tabular}{lcccccc}
\toprule
\multirow{2}*{methods} & \multicolumn{3}{c}{Cora}  & \multicolumn{3}{c}{Amazon-Photo} \\
\cmidrule(lr){2-4}\cmidrule(lr){5-7}
& GCN& GAT & SGC& GCN& GAT & SGC\\ 
% \cline{2-3}
\toprule

ERM
&67.28\tiny{$\pm$5.37}
&72.39\tiny{$\pm$6.45} 
&64.79\tiny{$\pm$4.78}
&88.06\tiny{$\pm$0.65}
&75.85\tiny{$\pm$1.23} 
&79.26\tiny{$\pm$2.41}

 \\
EERM  & 
71.02\tiny{$\pm$4.52} & 
74.31\tiny{$\pm$4.87} &
66.21\tiny{$\pm$5.68} &
90.79\tiny{$\pm$2.18} &
81.14\tiny{$\pm$0.76} &
81.56\tiny{$\pm$1.53} 

 \\

InfoIGL  & 
\textbf{72.28\tiny{$\pm$4.86}}$\uparrow$& 
\textbf{77.40\tiny{$\pm$6.36}}$\uparrow$& 
\textbf{69.08\tiny{$\pm$4.47}}$\uparrow$&
\textbf{91.73\tiny{$\pm$0.47}}$\uparrow$&
\textbf{82.03\tiny{$\pm$2.02}}$\uparrow$&
\textbf{83.28\tiny{$\pm$ 1.05}}$\uparrow$
 \\
 % \hline

 \hline

\end{tabular}}
\end{table*}

\subsection{Scalability Analysis (Q5)}
To extend our method on solving the OOD problems of node classification tasks, we follow the setting of EERM~\cite{wu2022handling} and implement InfoIGL on the dataset of Cora and Amazon-Photo. The performance of InfoIGL on node classification tasks is listed in Table~\ref{table:node}. InfoIGL outperforms ERM and EERM~\cite{wu2022handling} on both of Cora and Amazon-photo datasets, achieving notable performance improvements on different backbones ( including GCN, GAT, and SGC).

\subsection{Visualization of the Invariance}
To validate the effectiveness of our method in extracting invariant features from graphs of the same class, we visually examine a selection of randomly sampled graphs during training to illustrate the instantiated invariance obtained by InfoIGL. We employ a GIN-based encoder and apply InfoIGL specifically on motif-size graphs. The visualizations are presented in Figure \ref{fig:visual}. The original graphs are depicted on the left side of the grids, where the green regions represent motifs (such as a house, crane, or cycle) that remain invariant across graphs within the same class. Conversely, the yellow portions indicate the base graph (such as a wheel, tree, star, or path) whose size may cause a distribution shift. The ground-truths of invariance are labeled by the dataset creator. On the right side of the grids, we display the identified invariance after training using InfoIGL. We colorize nodes and edges based on the attention weights assigned by the model, and the nodes with darker colors indicate a higher degree of invariance. Remarkably, we observe that almost all of the darker colors accurately align with the invariant areas. This further demonstrates the efficacy of InfoIGL in successfully capturing invariant features through information bottleneck theory.

\subsection{Running Time Comparison}
In our experiment, we conduct a comparative analysis of the running time between ERM, InfoIGL, and its variants (including InfoIGL-S and InfoIGL-I) on various datasets. The results are detailed in Table~\ref{table:time}. We can observe that the time spent by InfoIGL is close to that of ERM, with a slight increase attributed to contrastive learning. This observation highlights the reasonable time complexity and efficiency of InfoIGL. Moreover, thanks to InfoIGL's exceptional capability in tackling graph OOD problems, we can circumvent the need for retraining or finetuning the model under varied data distributions, which can conserve computational resources and enhance both time and space efficiency.

\begin{table*}[htbp]
\renewcommand\arraystretch{1.2}
\centering
\caption{Comparison of running time for ERM, InfoIGL, and its variants.  }
\vspace{2mm}
\label{table:time}
\setlength{\tabcolsep}{4mm}{\begin{tabular}{lccccc}
\toprule
Dataset & ERM &InfoIGL-S&InfoIGL-I&InfoIGL\\

% & GCN& GAT & SGC& GCN& GAT & SGC\\ 
% \cline{2-3}
\toprule
Motif-size  & 
00h 21m 32s &
00h 20m 41s&
00h 20m 33s&
00h 21m 21s
 \\

Motif-base  & 
00h 33m 10s &
00h 32m 29s&
00h 32m 48s&
00h 33m 49s  
 \\
 
HIV-size&
00h 08m 43s&
00h 10m 44s&
00h 12m 07s&
00h 12m 39s

 \\
HIV-scaffold  & 
00h 14m 27s &
00h 15m 08s &
00h 14m 23s&
00h 16m 13s
 \\

Molbbbp-size  & 
00h 03m 44s&
00h 03m 31s&
00h 03m 54s&
00h 04m 29s

 \\

Molbbbp-scaffold  & 
00h 03m 45s&
00h 06m 18s&
00h 06m 37s&
00h 06m 37s
 \\
CMNIST-color&
00h 49m 56s &
00h 55m 58s&
00h 59m 22s &
01h 06m 36s
\\

 \hline

\end{tabular}}
\end{table*}

\section{Limitations}
\textbf{Informal hard negative mining. }There are several techniques for generating hard negative samples in machine learning. One approach is to choose negative samples that closely resemble positive examples by sampling from a pool of negatives. These selected samples can be relabeled as ``hard negatives'' and included in the training process. Another method involves the use of sophisticated algorithms like online hard example mining (OHEM), which identifies challenging negative samples based on their loss values during training. However, instead of these methods, we select hard negative samples by computing the distance between the negative samples and the semantic center that corresponds to the positive sample. While this informal hard negative mining technique may conserve computational resources, it could also introduce a certain degree of error.

\textbf{Lack of testing on larger real-world datasets.}
Due to the scarcity of mature datasets in the field of graph OOD generalization, we initially validated the effectiveness of our approach on the dataset of “GOOD”~\cite{gui2022good}, including Motif, HIV, CMNIST, and Molbbbp. These datasets are created with distribution shifts, which can be divided into two kinds: concept shift and covariate shift. We plan to further validate InfoIGL on a larger and more comprehensive dataset as the next step. Besides, we have the utmost confidence in the effectiveness of our method for performing cross-domain fusion when the dataset quality is ensured. In the future, we will explore expanding the method to tasks like analyzing material structures, predicting properties, or understanding material relationships. The applicability of InfoIGL on larger real-world datasets is what we leave for future work.

% \subsection{Potential Negative Social Impacts}
% As our framework can extract the invariant representations with information theory and realize reliable GNNs which can alleviate graph OOD problems, we are confident that the benefits of our work outweigh any potential negative impacts. However, the trustworthy GNN may cause over-reliance and be widely applied in the real world, leading to decreased human ability and consequent unemployment. Additionally, excessive trust in this technology may cause significant losses due to potential error overlooking. Therefore, it is crucial that people use the framework prudently and increase supervision when applying it.

%% file: 6_conclusion.tex
\section{Conclusion}
In this paper, we propose a novel framework, InfoIGL, to extract invariant representation for graph OOD generalization based on information bottleneck theory. To satisfy the invariance and sufficiency conditions of invariant learning, we compress redundant information from spurious features with redundancy filter and maximize mutual information of graphs from the same class with multi-level contrastive learning. Extensive experiments demonstrate the superiority of InfoIGL, highlighting its potential for real-world applications.

\section{Acknowledgements}
This research is supported by the National Natural Science Foundation of China (No.92270114 and No.62302321) and the advanced computing resources provided by the Supercomputing Center of the USTC.

%% file: 7_appendix.tex
{
\section{Notation}
For a better understanding, We summarize and provide some key notations in Table \ref{table: notation}.
\begin{table}[htbp]
\renewcommand{\arraystretch}{0.5}
\centering
\caption{Notations used in our paper.}
\label{table: notation}
{\begin{tabular}{p{1cm}p{6.5cm}}
\toprule
Notation & Explanation \\
% & GCN& GAT & SGC& GCN& GAT & SGC\\ 
% \cline{2-3}
\toprule
$\mathbb{G}$ &The graph space \\
$\mathbb{Y}$ &The label space \\
$\mathbf{G}$ & Random variable for the graph sample\\
$\mathbf{X}$ &The node feature matrix for graph $\mathbf{G}$\\
$\mathbf{A}$ &The adjacent matrix for graph $\mathbf{G}$\\
$\mathbf{Y}$ & Random variable of the label for graph $\mathbf{G}$\\
$\mathcal{D}$ &The dataset \\
 % $\mathcal{D}_{\text{te}}$ &  \\
$\mathcal{E}$ &The environment set \\
 % $\mathcal{E}_{\text{te}}$ &  \\
$\mathbf{e}$ &The environment factor \\
$\mathbf{\Phi}$ &The graph encoder \\
$\mathbf{\theta}$ &The classifier \\
 %$f$ & The graph model \\
$\mathbf{h}$ & The graph representation after GNN \\
$\mathbf{z}$ &The latent representation after projection head \\
$\mathbf{w}$ &The semantic representation \\
$\mathbf{h}_{G},$ &The GNN's embedding of the graph $G$\\
$\mathbf{z}_{G},$ &The latent representations of the graph $G$ after projection head\\
$G_+, \mathbf{z}_{G_+}$ &The positive samples of the graphs and the corresponding representations\\
$G_-,\mathbf{z}_{G_-}$ &The negative samples of the graphs and the corresponding representations\\
$\mathbf{z}_{inv}$ &The invariant features \\
$\mathbf{z}_{sup}$ &The spurious features \\
$\mathbf{h}_u^{(k)}, \mathbf{h}_{uv}^{(k)}$ & The node and edge embedding obtained from the $k^{th}$ layer of GNN \\
$\mathbf{w}_c$ & The semantic
representation over examples belonging to class $c$ \\
$\lambda_c, \lambda_i, \lambda_s$ & The non-negative weight for instance constraint, instance- and semantic-level contrastive loss\\

 \hline

\end{tabular}}
\end{table}
}

%% file: main.bbl
\begin{thebibliography}{10}

\bibitem{FCS_wu}
Wu~J, He~X, Wang X, Wang Q, Chen W, Lian J, Xie X.
\newblock Graph convolution machine for context-aware recommender system.
\newblock Frontiers Comput. Sci., 2022, 16(6): 166614

\bibitem{RPP_TOIS}
Mao W, Wu~J, Chen W, Gao C, Wang X, He~X.
\newblock Reinforced prompt personalization for recommendation with large language models.
\newblock CoRR, 2024, abs/2407.17115

\bibitem{mao2025distinguishedquantizedguidancediffusionbased}
Mao W, Liu S, Liu H, Liu H, Li~X, Hu~L.
\newblock Distinguished quantized guidance for diffusion-based sequence recommendation.
\newblock In: The Web Conference.
\newblock 2025

\bibitem{tois_Wu}
Wu~J, Wang X, Gao X, Chen J, Fu~H, Qiu T.
\newblock On the effectiveness of sampled softmax loss for item recommendation.
\newblock {ACM} Transactions on Information Systems, 2024, 42(4): 98:1--98:26

\bibitem{WWW_GIF}
Wu~J, Yang Y, Qian Y, Sui Y, Wang X, He~X.
\newblock {GIF:} {A} general graph unlearning strategy via influence function.
\newblock In: The Web Conference.
\newblock 2023,  651--661

\bibitem{xu2018powerful}
Xu~K, Hu~W, Leskovec J, Jegelka S.
\newblock How powerful are graph neural networks?
\newblock In: International Conference on Learning Representations.
\newblock 2019

\bibitem{tkdd_ood}
Sui Y, Mao W, Wang S, Wang X, Wu~J, He~X, Chua T.
\newblock Enhancing out-of-distribution generalization on graphs via causal attention learning.
\newblock ACM Transactions on Knowledge Discovery from Data, 2024, 18(5): 127:1--127:24

\bibitem{wang2022unified}
Wang Z, Veitch V.
\newblock A unified causal view of domain invariant representation learning.
\newblock In: ICML 2022: Workshop on Spurious Correlations, Invariance and Stability.
\newblock 2022

\bibitem{DIR}
Wu~Y, Wang X, Zhang A, He~X, Chua T~S.
\newblock Discovering invariant rationales for graph neural networks.
\newblock In: International Conference on Learning Representations.
\newblock 2022

\bibitem{arjovsky2019invariant}
Arjovsky M, Bottou L, Gulrajani I, Lopez-Paz D.
\newblock Invariant risk minimization.
\newblock arXiv preprint arXiv:1907.02893, 2019

\bibitem{wu2022handling}
Wu~Q, Zhang H, Yan J, Wipf D.
\newblock Handling distribution shifts on graphs: An invariance perspective.
\newblock In: {International Conference on Learning Representations}.
\newblock 2022

\bibitem{liu2022graph}
Liu G, Zhao T, Xu~J, Luo T, Jiang M.
\newblock Graph rationalization with environment-based augmentations.
\newblock In: ACM SIGKDD Conference on Knowledge Discovery and Data Mining.
\newblock 2022,  1069--1078

\bibitem{li2022learning}
Li~H, Zhang Z, Wang X, Zhu W.
\newblock Learning invariant graph representations for out-of-distribution generalization.
\newblock In: Advances in Neural Information Processing Systems.
\newblock 2022,  11828--11841

\bibitem{wang2021mixup}
Wang Y, Wang W, Liang Y, Cai Y, Hooi B.
\newblock Mixup for node and graph classification.
\newblock In: The Web Conference.
\newblock 2021,  3663--3674

\bibitem{fan2021generalizing}
Fan S, Wang X, Shi C, Cui P, Wang B.
\newblock Generalizing graph neural networks on out-of-distribution graphs.
\newblock IEEE Transactions on Pattern Analysis and Machine Intelligence, 2024, 46(1): 322--337

\bibitem{li2021ood}
Li~H, Wang X, Zhang Z, Zhu W.
\newblock {OOD-GNN:} out-of-distribution generalized graph neural network.
\newblock IEEE Transactions on Knowledge and Data Engineering, 2023, 35(7): 7328--7340

\bibitem{sui2022causal}
Sui Y, Wang X, Wu~J, Lin M, He~X, Chua T.
\newblock Causal attention for interpretable and generalizable graph classification.
\newblock In: {ACM SIGKDD Conference on Knowledge Discovery and Data Mining}.
\newblock 2022,  1696--1705

\bibitem{pearl2014interpretation}
Pearl J.
\newblock Interpretation and identification of causal mediation.
\newblock Psychological methods, 2014, 19(4): 459

\bibitem{saxe2019information}
Saxe A~M, Bansal Y, Dapello J, Advani M, Kolchinsky A, Tracey B~D, Cox D~D.
\newblock On the information bottleneck theory of deep learning.
\newblock In: {International Conference on Learning Representations} (Poster).
\newblock 2018

\bibitem{yue2021prototypical}
Yue X, Zheng Z, Zhang S, Gao Y, Darrell T, Keutzer K, Sangiovanni{-}Vincentelli A~L.
\newblock Prototypical cross-domain self-supervised learning for few-shot unsupervised domain adaptation.
\newblock In: {Conference on Computer Vision and Pattern Recognition}.
\newblock 2021,  13834--13844

\bibitem{wang2022cross}
Wang R, Wu~Z, Weng Z, Chen J, Qi~G, Jiang Y.
\newblock Cross-domain contrastive learning for unsupervised domain adaptation.
\newblock {IEEE} Trans. Multim., 2023, 25: 1665--1673

\bibitem{yao2022pcl}
Yao X, Bai Y, Zhang X, Zhang Y, Sun Q, Chen R, Li~R, Yu~B.
\newblock {PCL:} proxy-based contrastive learning for domain generalization.
\newblock In: {Conference on Computer Vision and Pattern Recognition}.
\newblock 2022,  7087--7097

\bibitem{zhao2019learning}
Zhao H, Combes d~R~T, Zhang K, Gordon G~J.
\newblock On learning invariant representations for domain adaptation.
\newblock In: {International Conference on Machine Learning}.
\newblock 2019,  7523--7532

\bibitem{rosenfeld2020risks}
Rosenfeld E, Ravikumar P~K, Risteski A.
\newblock The risks of invariant risk minimization.
\newblock In: International Conference on Learning Representations.
\newblock 2020

\bibitem{yang2020learning}
Yang S, Fu~K, Yang X, Lin Y, Zhang J, Cheng P.
\newblock Learning domain-invariant discriminative features for heterogeneous face recognition.
\newblock {IEEE} Access, 2020, 8: 209790--209801

\bibitem{miao2022interpretable}
Miao S, Liu M, Li~P.
\newblock Interpretable and generalizable graph learning via stochastic attention mechanism.
\newblock In: {International Conference on Machine Learning}.
\newblock 2022,  15524--15543

\bibitem{rong2019dropedge}
Rong Y, Huang W, Xu~T, Huang J.
\newblock Dropedge: Towards deep graph convolutional networks on node classification.
\newblock In: {International Conference on Learning Representations}.
\newblock 2020

\bibitem{krueger2021out}
Krueger D, Caballero E, Jacobsen J~H, Zhang A, Binas J, Zhang D, Priol R~L, Courville A.
\newblock Out-of-distribution generalization via risk extrapolation (rex).
\newblock In: Proceedings of the 38th International Conference on Machine Learning.
\newblock 18--24 Jul 2021,  5815--5826

\bibitem{sagawa2019distributionally}
Sagawa S, Koh P~W, Hashimoto T~B, Liang P.
\newblock Distributionally robust neural networks for group shifts: On the importance of regularization for worst-case generalization.
\newblock In: International Conference on Learning Representations.
\newblock 2020

\bibitem{kong2022robust}
Kong K, Li~G, Ding M, Wu~Z, Zhu C, Ghanem B, Taylor G, Goldstein T.
\newblock Robust optimization as data augmentation for large-scale graphs.
\newblock In: {Conference on Computer Vision and Pattern Recognition}.
\newblock 2022,  60--69

\bibitem{federici2021information}
Federici M, Tomioka R, Forr{\'{e}} P.
\newblock An information-theoretic approach to distribution shifts.
\newblock In: Advances in Neural Information Processing Systems.
\newblock 2021,  17628--17641

\bibitem{fan2022debiasing}
Fan S, Wang X, Mo~Y, Shi C, Tang J.
\newblock Debiasing graph neural networks via learning disentangled causal substructure.
\newblock In: Advances in Neural Information Processing Systems.
\newblock 2022,  24934--24946

\bibitem{sui2024unleashing}
Sui Y, Wu~Q, Wu~J, Cui Q, Li~L, Zhou J, Wang X, He~X.
\newblock Unleashing the power of graph data augmentation on covariate distribution shift.
\newblock In: Advances in Neural Information Processing Systems.
\newblock 2023,  18109--18131

\bibitem{yang2022learning}
Yang N, Zeng K, Wu~Q, Jia X, Yan J.
\newblock Learning substructure invariance for out-of-distribution molecular representations.
\newblock In: Advances in Neural Information Processing Systems.
\newblock 2022,  12964--12978

\bibitem{gui2024joint}
Gui S, Liu M, Li~X, Luo Y, Ji~S.
\newblock Joint learning of label and environment causal independence for graph out-of-distribution generalization.
\newblock In: Advances in Neural Information Processing Systems.
\newblock 2023,  3945--3978

\bibitem{zhuang2024learning}
Zhuang X, Zhang Q, Ding K, Bian Y, Wang X, Lv~J, Chen H, Chen H.
\newblock Learning invariant molecular representation in latent discrete space.
\newblock In: Advances in Neural Information Processing Systems.
\newblock 2023,  78435--78452

\bibitem{han2022g}
Han X, Jiang Z, Liu N, Hu~X.
\newblock G-mixup: Graph data augmentation for graph classification.
\newblock In: {International Conference on Machine Learning}.
\newblock 2022,  8230--8248

\bibitem{chen2024does}
Chen Y, Bian Y, Zhou K, Xie B, Han B, Cheng J.
\newblock Does invariant graph learning via environment augmentation learn invariance?
\newblock In: Advances in Neural Information Processing Systems.
\newblock 2023,  71486--71519

\bibitem{tishby2000information}
Tishby N, Pereira F~C, Bialek W.
\newblock The information bottleneck method.
\newblock arXiv preprint physics/0004057, 2000

\bibitem{fang2024regularization}
Fang J, Zhang G, Wang K, Du~W, Duan Y, Wu~Y, Zimmermann R, Chu X, Liang Y.
\newblock On regularization for explaining graph neural networks: An information theory perspective.
\newblock IEEE Transactions on Knowledge and Data Engineering, 2024,  1--14

\bibitem{ye2022ood}
Ye~N, Li~K, Bai H, Yu~R, Hong L, Zhou F, Li~Z, Zhu J.
\newblock Ood-bench: Quantifying and understanding two dimensions of out-of-distribution generalization.
\newblock In: {Conference on Computer Vision and Pattern Recognition}.
\newblock 2022,  7937--7948

\bibitem{du2020learning}
Du~Y, Xu~J, Xiong H, Qiu Q, Zhen X, Snoek C~G~M, Shao L.
\newblock Learning to learn with variational information bottleneck for domain generalization.
\newblock In: {European Conference on Computer Vision}.
\newblock 2020,  200--216

\bibitem{ahuja2021invariance}
Ahuja K, Caballero E, Zhang D, Gagnon{-}Audet J, Bengio Y, Mitliagkas I, Rish I.
\newblock Invariance principle meets information bottleneck for out-of-distribution generalization.
\newblock In: Advances in Neural Information Processing Systems.
\newblock 2021,  3438--3450

\bibitem{li2022invariant}
Li~B, Shen Y, Wang Y, Zhu W, Reed C, Li~D, Keutzer K, Zhao H.
\newblock Invariant information bottleneck for domain generalization.
\newblock In: {AAAI}.
\newblock 2022,  7399--7407

\bibitem{poole2019variational}
Poole B, Ozair S, Oord v.~d A, Alemi A~A, Tucker G.
\newblock On variational bounds of mutual information.
\newblock In: {International Conference on Machine Learning}.
\newblock 2019,  5171--5180

\bibitem{he2020momentum}
He~K, Fan H, Wu~Y, Xie S, Girshick R~B.
\newblock Momentum contrast for unsupervised visual representation learning.
\newblock In: {Conference on Computer Vision and Pattern Recognition}.
\newblock 2020,  9726--9735

\bibitem{chen2020simple}
Chen T, Kornblith S, Norouzi M, Hinton G~E.
\newblock A simple framework for contrastive learning of visual representations.
\newblock In: {International Conference on Machine Learning}.
\newblock 2020,  1597--1607

\bibitem{khosla2020supervised}
Khosla P, Teterwak P, Wang C, Sarna A, Tian Y, Isola P, Maschinot A, Liu C, Krishnan D.
\newblock Supervised contrastive learning.
\newblock In: Advances in Neural Information Processing Systems.
\newblock 2020,  18661--18673

\bibitem{tian2020makes}
Tian Y, Sun C, Poole B, Krishnan D, Schmid C, Isola P.
\newblock What makes for good views for contrastive learning?
\newblock In: Advances in Neural Information Processing Systems.
\newblock 2020,  6827--6839

\bibitem{hassani2020contrastive}
Hassani K, Ahmadi A~H~K.
\newblock Contrastive multi-view representation learning on graphs.
\newblock In: {International Conference on Machine Learning}.
\newblock 2020,  4116--4126

\bibitem{huang2021towards}
Huang W, Yi~M, Zhao X, Jiang Z.
\newblock Towards the generalization of contrastive self-supervised learning.
\newblock In: {International Conference on Learning Representations}.
\newblock 2023

\bibitem{zhang2022correct}
Zhang M, Sohoni N~S, Zhang H~R, Finn C, Re~C.
\newblock Correct-n-contrast: a contrastive approach for improving robustness to spurious correlations.
\newblock In: International Conference on Machine Learning.
\newblock 2022,  26484--26516

\bibitem{chenlearning}
Chen Y, Zhang Y, Bian Y, Yang H, Kaili M, Xie B, Liu T, Han B, Cheng J.
\newblock Learning causally invariant representations for out-of-distribution generalization on graphs.
\newblock In: Advances in Neural Information Processing Systems.
\newblock 2022,  22131--22148

\bibitem{Auxiliaryood}
Boonlia H, Dam T, Ferdaus M~M, Anavatti S~G, Mullick A.
\newblock Improving self-supervised learning for out-of-distribution task via auxiliary classifier.
\newblock 2022 IEEE International Conference on Image Processing (ICIP), 2022,  3036--3040

\bibitem{yang2024invariant}
Yang M, Fang Z, Zhang Y, Du~Y, Liu F, Ton J~F, Wang J, Wang J.
\newblock Invariant learning via probability of sufficient and necessary causes.
\newblock In: Advances in Neural Information Processing Systems.
\newblock 2023,  79832--79857

\bibitem{xu2021self}
Xu~M, Wang H, Ni~B, Guo H, Tang J.
\newblock Self-supervised graph-level representation learning with local and global structure.
\newblock In: {International Conference on Machine Learning}.
\newblock 2021,  11548--11558

\bibitem{zhang2022leverage}
Zhang T, Qiu C, Ke~W, S{\"{u}}sstrunk S, Salzmann M.
\newblock Leverage your local and global representations: {A} new self-supervised learning strategy.
\newblock In: {Conference on Computer Vision and Pattern Recognition}.
\newblock 2022,  16559--16568

\bibitem{brody2021attentive}
Brody S, Alon U, Yahav E.
\newblock How attentive are graph attention networks?
\newblock In: International Conference on Learning Representations.
\newblock 2022

\bibitem{mutualesti}
Belghazi M~I, Baratin A, Rajeswar S, Ozair S, Bengio Y, Hjelm R~D, Courville A~C.
\newblock Mutual information neural estimation.
\newblock In: {International Conference on Machine Learning}.
\newblock 2018,  530--539

\bibitem{jing2021understanding}
Jing L, Vincent P, LeCun Y, Tian Y.
\newblock Understanding dimensional collapse in contrastive self-supervised learning.
\newblock In: {International Conference on Learning Representations}.
\newblock 2022

\bibitem{xuan2020hard}
Xuan H, Stylianou A, Liu X, Pless R.
\newblock Hard negative examples are hard, but useful.
\newblock In: {European Conference on Computer Vision}.
\newblock 2020,  126--142

\bibitem{robinson2020contrastive}
Robinson J~D, Chuang C, Sra S, Jegelka S.
\newblock Contrastive learning with hard negative samples.
\newblock In: {International Conference on Learning Representations}.
\newblock 2021

\bibitem{gui2022good}
Gui S, Li~X, Wang L, Ji~S.
\newblock Good: A graph out-of-distribution benchmark.
\newblock In: Advances in Neural Information Processing Systems.
\newblock 2022,  2059--2073

\bibitem{yang2016revisiting}
Yang Z, Cohen W, Salakhudinov R.
\newblock Revisiting semi-supervised learning with graph embeddings.
\newblock In: International Conference on Machine Learning.
\newblock 2016,  40--48

\bibitem{rozemberczki2021multi}
Rozemberczki B, Allen C, Sarkar R.
\newblock Multi-scale attributed node embedding.
\newblock J. Complex Networks, 2021, 9(2)

\bibitem{ying2019gnnexplainer}
Ying Z, Bourgeois D, You J, Zitnik M, Leskovec J.
\newblock Gnnexplainer: Generating explanations for graph neural networks.
\newblock In: Advances in Neural Information Processing Systems.
\newblock 2019,  9240--9251

\bibitem{hu2020open}
Hu~W, Fey M, Zitnik M, Dong Y, Ren H, Liu B, Catasta M, Leskovec J.
\newblock Open graph benchmark: Datasets for machine learning on graphs.
\newblock In: Advances in Neural Information Processing Systems.
\newblock 2020,  22118--22133

\bibitem{wu2018moleculenet}
Wu~Z, Ramsundar B, Feinberg E~N, Gomes J, Geniesse C, Pappu A~S, Leswing K, Pande V.
\newblock Moleculenet: a benchmark for molecular machine learning.
\newblock Chemical science, 2018, 9(2): 513--530

\bibitem{peng2020graph}
Peng Z, Huang W, Luo M, Zheng Q, Rong Y, Xu~T, Huang J.
\newblock Graph representation learning via graphical mutual information maximization.
\newblock In: The Web Conference.
\newblock 2020,  259--270

\bibitem{sun2019infograph}
Sun F, Hoffmann J, Verma V, Tang J.
\newblock Infograph: Unsupervised and semi-supervised graph-level representation learning via mutual information maximization.
\newblock In: {International Conference on Learning Representations}.
\newblock 2020

\bibitem{hafidi2007graphcl}
Hafidi H, Ghogho M, Ciblat P, Swami A.
\newblock Graphcl: Contrastive self-supervised learning of graph representations.
\newblock CoRR, 2020, abs/2007.08025

\end{thebibliography}
